\documentclass[11pt]{article}

\usepackage[final]{acl}

\usepackage{times}
\usepackage{latexsym}
\usepackage[T1]{fontenc}
\usepackage[utf8]{inputenc}
\usepackage{microtype}
\usepackage{inconsolata}
\usepackage{graphicx}
\usepackage{amsfonts}
\usepackage{amsmath}
\usepackage{multirow}
\usepackage{booktabs}
\usepackage[table]{xcolor}
\newcommand{\second}[1]{\underline{#1}}

\title{Stable Language Guidance for Vision–Language–Action Models}

\author{
    \textbf{Zhihao Zhan\textsuperscript{1}},
    \textbf{Yuhao Chen\textsuperscript{1}},
    \textbf{Jiaying Zhou\textsuperscript{1}},
    \textbf{Qinhan Lyu\textsuperscript{1}},
    \textbf{Hao Liu\textsuperscript{1}},\\
    \textbf{Keze Wang\textsuperscript{1,2,3}},
    \textbf{Liang Lin\textsuperscript{1,2,3}},
    \textbf{Guangrun Wang \textsuperscript{1,2,3}\thanks{Corresponding author.}}\\[0.5em]
    \textsuperscript{1}Sun Yat-sen University\\
    \textsuperscript{2}Guangdong Key Lab of Big Data Analysis \& Processing\\
    \textsuperscript{3}X-Era AI Lab\\[0.5em]
    \small{
        \texttt{zhanzhh6@mail2.sysu.edu.cn, wanggrun@gmail.com}
    }
}

\begin{document}
\maketitle

\begin{abstract}
Vision-Language-Action (VLA) models have demonstrated impressive capabilities in generalized robotic control; however, they remain notoriously brittle to linguistic perturbations. We identify a critical ``modality collapse'' phenomenon where strong visual priors overwhelm sparse linguistic signals, causing agents to overfit to specific instruction phrasings while ignoring the underlying semantic intent. To address this, we propose \textbf{Residual Semantic Steering (RSS)}, a probabilistic framework that disentangles physical affordance from semantic execution. RSS introduces two theoretical innovations: (1) \textbf{Monte Carlo Syntactic Integration}, which approximates the true semantic posterior via dense, LLM-driven distributional expansion, and (2) \textbf{Residual Affordance Steering}, a dual-stream decoding mechanism that explicitly isolates the causal influence of language by subtracting the visual affordance prior. Theoretical analysis suggests that RSS effectively maximizes the mutual information between action and intent while suppressing visual distractors. Empirical results across diverse manipulation benchmarks demonstrate that RSS achieves state-of-the-art robustness, maintaining performance even under adversarial linguistic perturbations. We release our code at \url{https://github.com/Doo-mon/RSS}.
\end{abstract}
\section{Introduction} \label{sec:intro}

The core promise of Embodied AI is the ability to map high-level intent to low-level control. Formally, we seek a policy $\pi(a | o, z)$ where $o$ is the observation and $z$ is the latent semantic intent. However, in practice, we only have access to linguistic realizations $l \sim p(l|z)$. Current Vision-Language-Action (VLA) models approximate $\pi(a | o, l)$, but often learn a degenerate mapping where $\pi(a | o, l_i) \not\approx \pi(a | o, l_j)$ even when $l_i$ and $l_j$ share the same intent $z$.

\begin{figure}[t]
  \centering
  \includegraphics[width=0.5\textwidth]{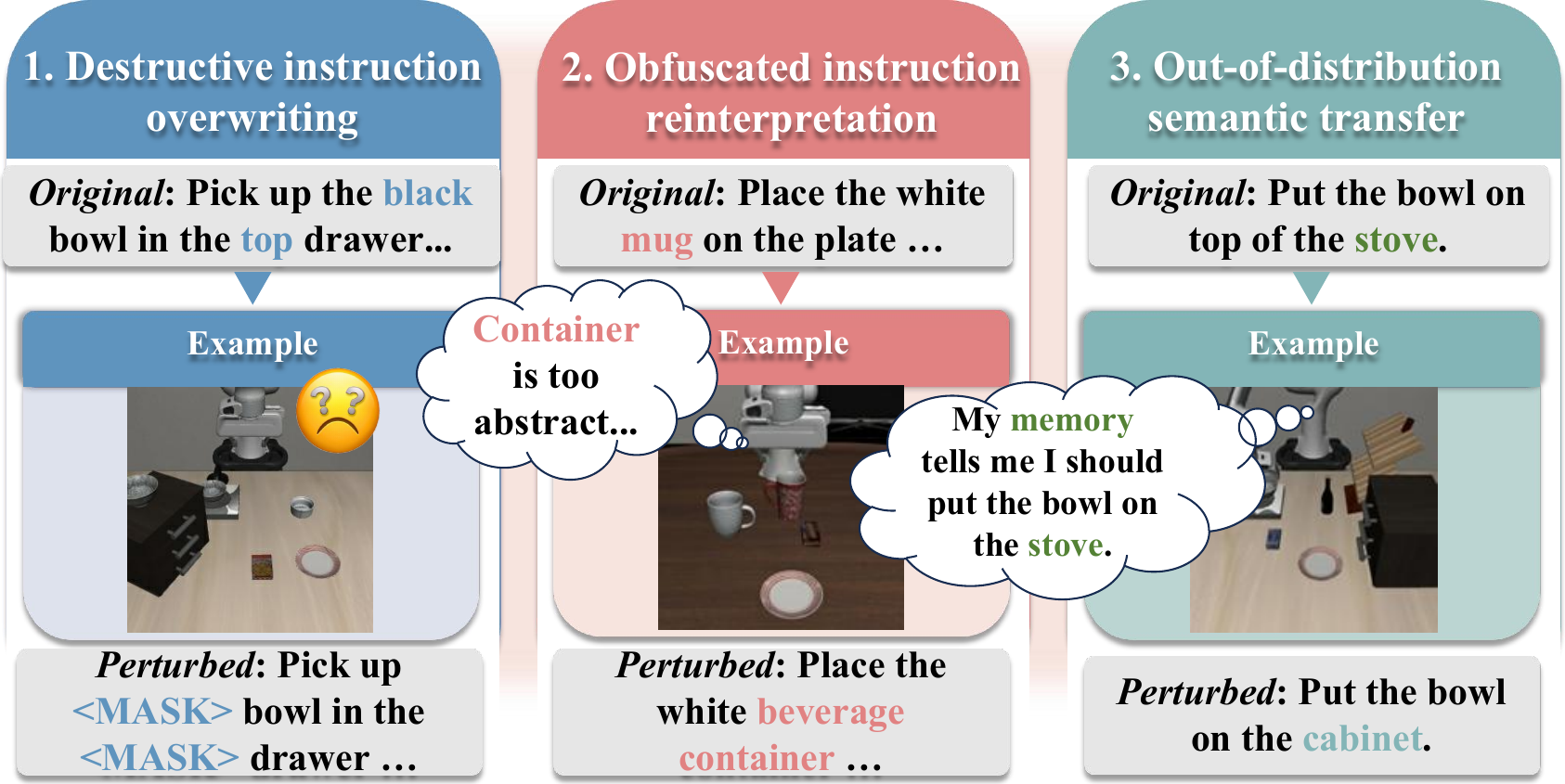}
  \caption{\textbf{Taxonomy of Language Instruction Perturbations.} We identify three distinct failure modes in VLA instruction following: (1) \textbf{Destructive Instruction Overwriting}, where critical semantic tokens are lost or masked (e.g., masking the drawer location); (2) \textbf{Obfuscated Instruction Reinterpretation}, where the model fails to ground synonymous or verbose descriptions (e.g., ``beverage container'' vs. ``mug''); and (3) \textbf{Out-of-Distribution Semantic Transfer}, where the instruction targets a valid but unlearned goal configuration (e.g., placing on a ``cabinet'' instead of the training-set ``stove'').}\label{fig:perturb_examples}
\end{figure}

Recent empirical audits validate this failure mode. Analyses on the \textbf{Libero-Plus} \citep{fei2025libero} and \textbf{RADAR} \citep{chen2026radar} benchmarks reveal a pervasive ``instruction blindness,'' where models frequently ignore language inputs entirely, defaulting to the most probable action given the scene alone. Furthermore, evidence from \textbf{Libero-Pro} highlights a critical lack of grounding; even when language is processed, models exhibit an overdependence on rote pattern execution rather than true semantic interpretation, leading to catastrophic failure when instructions deviate from training templates~\citep{zhou2025libero}. As illustrated in Figure~\ref{fig:perturb_examples}, these failures manifest through three primary categories of linguistic variation: destructive instruction overwriting, obfuscated reinterpretation of synonyms, and out-of-distribution semantic transfer.

We argue that this fragility stems from two sources: \begin{enumerate}  \item \textbf{Manifold Sparsity:} The training data covers a vanishingly small support of the syntactic distribution $p(l|z)$, leading to overfitting on surface statistics.    \item \textbf{Prior Dominance:} In the high-dimensional joint distribution $p(a|o, l)$, the visual signal $o$ contains dense, high-frequency information (edges, textures) that dominates the gradient, causing the model to default to a ``visual affordance prior'' (e.g., ``grasp the nearest object'') regardless of the text.\end{enumerate}

We introduce \textbf{Residual Semantic Steering (RSS)}, a framework that addresses these challenges through two theoretical innovations. First, to resolve manifold sparsity, we employ \textbf{Monte Carlo Syntactic Integration}. Instead of relying on single data points, we treat the instruction as a seed and utilize an Oracle Teacher (LLM) to generate a dense syntactic neighborhood. By optimizing an Expected Semantic Loss over this expanded distribution, we force the policy to marginalize out syntactic noise and approximate the true posterior $\pi(a|o, z)$. Second, to counteract prior dominance, we introduce \textbf{Residual Affordance Steering}. We reinterpret the ``unconditional'' forward pass not as a null baseline, but as the Base Affordance Distribution $s(a|o, \emptyset)$—capturing physically feasible actions independent of intent. By subtracting this visual prior from the conditional logits, we isolate a Pure Semantic Signal that represents the causal influence of language. Unlike standard Classifier-Free Guidance (CFG) \citep{ho2022classifier}, which acts as a ``quality booster'' in generative models, RSS functions as a Bias Suppressor in control, mathematically penalizing actions driven solely by visual instinct.

\textbf{Contributions.} Our main contributions are summarized as follows:
\begin{itemize}
\item We introduce Monte Carlo Syntactic Integration, a training strategy that utilizes an Oracle Teacher to generate dense linguistic neighborhoods. By optimizing an Expected Semantic Loss over this expanded distribution, we approximate the true semantic posterior, ensuring the policy is invariant to surface-level syntactic perturbations.
\item We propose Residual Affordance Steering that acts as a Bias Suppressor. By subtracting the Base Affordance Distribution (visual prior) from the conditional logits, it isolates the Pure Semantic Signal, effectively restoring the rank of linguistic features relative to visual dominators.
\item We empirically demonstrate that RSS achieves state-of-the-art robustness, effectively mitigating ``instruction blindness'' and preventing rote pattern execution by decoupling semantic intent from visual affordances.
\end{itemize}

\section{Related Work} \label{sec:relate}

\paragraph{Vision-Language-Action Architectures.} 
The field has rapidly evolved from early large-scale imitators like RT-1~\citep{brohan2022rt} and RT-2~\citep{zitkovich2023rt} to efficient open-source models like OpenVLA~\citep{kim2024openvla}. Recent autoregressive approaches focus on specific capabilities: SpatialVLA~\citep{qu2025spatialvla} incorporates 3D spatial cues, OpenVLA-OFT~\citep{kim2025fine} optimizes continuous action tuning, and $\pi_0$ FAST~\citep{pertsch2025pi0fast} enhances training efficiency. Parallel work integrates chain-of-thought reasoning to guide planning (CoT-VLA~\citep{zhao2025cot}, GR-1~\citep{wu2023unleashing}). Simultaneously, diffusion-based control has matured from the seminal Diffusion Policy~\citep{chi2023diffusion} to scalable transformers like CogACT~\citep{li2024cogact} and RDT~\citep{liu2024rdt}. State-of-the-art generalist frameworks now employ flow matching ($\pi_0$~\citep{black2024pi_0}, $\pi_{0.5}$~\citep{intelligence2025pi05}), $\mathcal{E}_0$ \citep{zhan2026e0enhancinggeneralizationfinegrained}, dual-system designs (OneTwoVLA~\citep{lin2025onetwovla}, GR00T N1~\citep{bjorck2025gr00t}), and decoupling approaches \citep{li2025vla} to achieve broad physical generalization \citep{zhou2026tag,chen2026radar}.

\paragraph{Language Guidance and Modality Imbalance.}
Despite these architectural advances, balancing multi-modal inputs remains a critical bottleneck \citep{chen2026radar}. Recent audits on the \textbf{Libero-Plus}~\citep{fei2025libero} and \textbf{Libero-Pro}~\citep{zhou2025libero} benchmarks reveal that VLA agents frequently suffer from ``instruction blindness'' or rote execution, as dense high-frequency visual signals tend to overwhelm sparse linguistic tokens. A notable architectural solution is \textbf{RDT-1B}~\citep{liu2024rdt}, which mitigates this text overshadowing by eschewing simultaneous token injection. Instead, it employs an alternating cross-attention mechanism, injecting image and text tokens in successive layers. This strategy explicitly preserves the magnitude of linguistic gradients during deep fusion, ensuring that instruction signals are not drowned out by the visual affordance prior.

\section{Methodology: The RSS Framework}\label{sec:method}

\begin{figure*}[t]
  \centering
  \includegraphics[width=1.0\textwidth]{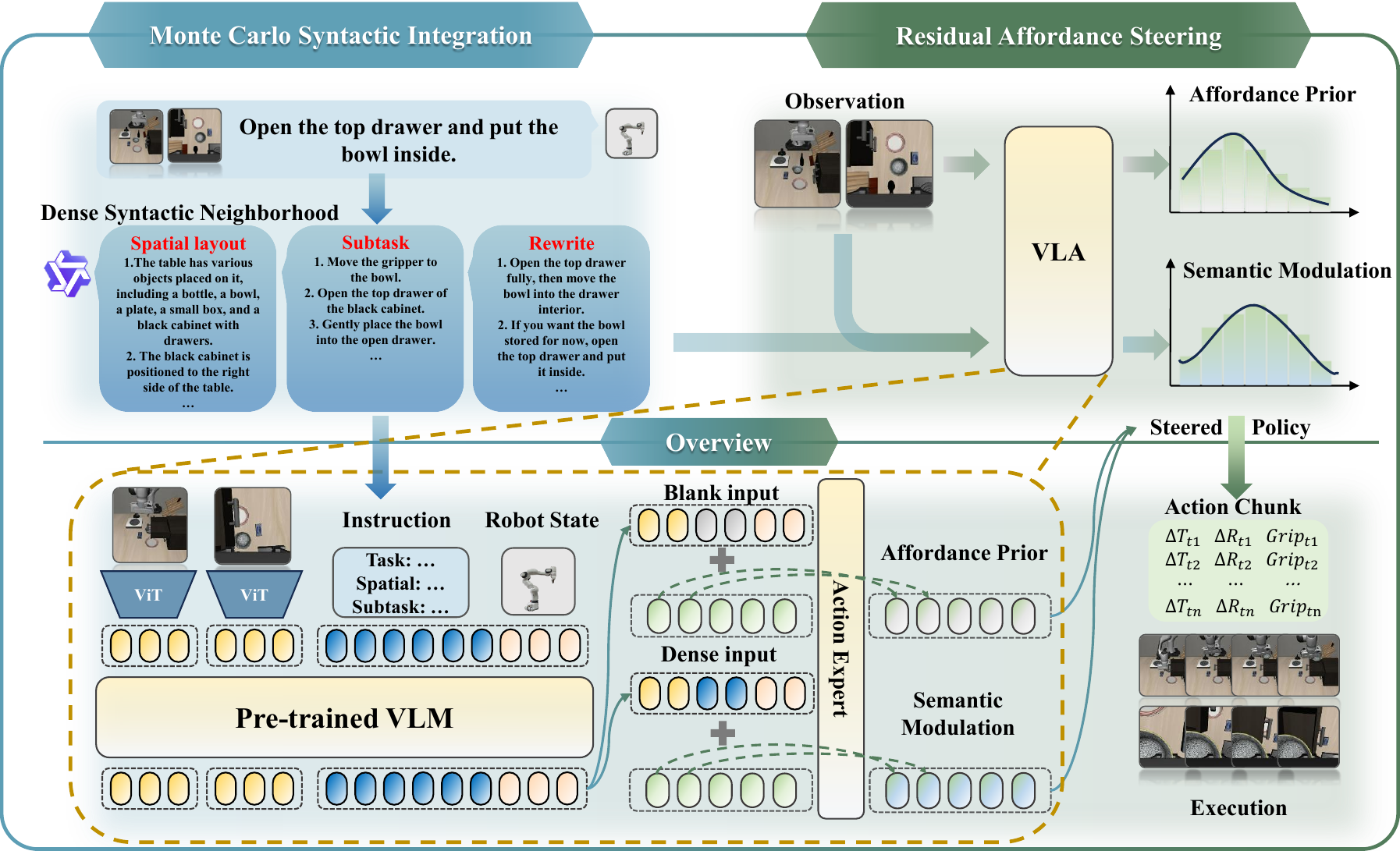}
  \caption{\textbf{Overview of Residual Semantic Steering (RSS).} 
To combat instruction blindness, RSS operates in two stages.
\textbf{Left:} \textit{Monte Carlo Syntactic Integration} utilizes an Oracle Teacher to generate a dense linguistic neighborhood around a seed instruction. Optimizing over this distribution forces the policy to learn representations that are invariant to syntactic perturbations.
\textbf{Right:} \textit{Residual Affordance Steering} mitigates visual prior dominance. By subtracting the unconditioned ``visual instinct'' (Base Affordance) from the standard prediction, we isolate and amplify the residual semantic signal, ensuring the policy follows the specific user intent rather than generic visual attractors.}\label{fig:main}
\end{figure*}

\subsection{Preliminaries}

Vision-language-action(VLA) models are commonly trained via imitation learning on large-scale robot demonstration datasets $\mathcal{D}$. 
Given an observation $o_t$ and a natural language task instruction $l$, the objective is to maximize the log-likelihood of an action (or an action chunk) $a_{t:t+H}$:
\begin{equation}
\max_{\theta}\; \mathbb{E}_{(a_{t:t+H},\, o_t,\, l)\sim\mathcal{D}}
\left[ \log \pi_{\theta}(a_{t:t+H} \mid o_t, l) \right].
\end{equation}
The observation $o_t$ typically comprises one or more visual inputs $\{I_t^1, \ldots, I_t^n\}$ and a proprioceptive state $q_t$ encoding the robot's joint configurations.

Modern VLA architectures inherit the design principles of large vision-language models, employing modality-specific tokenizers to map visual, linguistic, and action inputs into either discrete or continuous token representations \cite{brohan2022rt,zitkovich2023rt,pertsch2025pi0fast,black2024pi_0,intelligence2025pi05}. These tokens are processed by a large autoregressive transformer backbone, whose parameters are commonly initialized from pretrained vision-language models. By encoding observations, instructions, and actions into a unified token sequence, the imitation learning objective can be reformulated as a standard next-token prediction problem, enabling the use of scalable training techniques from contemporary language modeling.

We consider a VLA policy parameterized by $\theta$. Our goal is to ensure the policy is invariant to linguistic noise $\epsilon_l$ such that $\pi(a|o, l) \approx \pi(a|o, l + \epsilon_l)$.

To achieve this, we propose the \textbf{Residual Semantic Steering (RSS)} framework. As illustrated in Figure~\ref{fig:main}, RSS operates as a dual-stage mechanism: it densifies the linguistic supervision signal via \textit{Monte Carlo Syntactic Integration} to enforce semantic invariance; it actively suppresses visual priors via \textit{Residual Affordance Steering} to isolate the pure semantic intent.

\subsection{Monte Carlo Syntactic Integration}


Standard Maximum Likelihood Estimation (MLE) minimizes $\mathcal{L} = -\log p_\theta(a|o, l)$. However, a single instruction $l$ is a noisy estimator of the true intent $z$. For instance, consider the latent intent $z$ as ``grasping the apple.'' This intent can be realized through various instructions $l$, such as ``Pick up the red fruit,'' ``Fetch the apple,'' or ``Grab it.'' If the model is trained solely on ``Pick up the red fruit,'' it treats specific lexical choices (e.g., ``fruit'' vs. ``apple'') as an essential signal rather than what they truly are: syntactic noise partially obscuring the underlying semantic intent. To learn the true semantic policy $p(a|o, z)$, we must marginalize over the nuisance variable of syntax:\begin{equation}p(a|o, z) = \int p(a|o, l) p(l|z) \, dl\end{equation}Since this integral is intractable, we employ a \textbf{Monte Carlo approximation} using an Oracle Teacher (e.g., a high-capability LLM). We treat the original instruction $l_{orig}$ as a seed and generate a dense neighborhood $\mathcal{N}(l_{orig}) = \{l_1, ..., l_K\}$ via the Teacher, sampling from the induced distribution $\hat{p}(l|z)$.We then optimize the \textbf{Expected Semantic Loss}:\begin{equation}\mathcal{L}_{RSS} = \mathbb{E}_{(o, a) \sim \mathcal{D}} \left[ \frac{1}{K} \sum_{k=1}^K -\log \pi_\theta(a | o, l_k) \right]\end{equation}

This formulation forces the encoder to map disparate linguistic inputs $\{l_k\}$ to a unified region in the latent embedding space, explicitly minimizing the conditional entropy $H(A | L)$ with respect to syntactic variation.\

\subsection{Residual Affordance Steering}

Standard CFG combines conditional and unconditional scores. We re-interpret this for the action space to distinguish our method from generative diffusion. Let $s(a|o, l)$ be the logit score of an action. We decompose the decision process into two components:
\begin{enumerate}    
\item \textbf{The Affordance Prior $s(a|o, \emptyset)$:} The probability of an action based solely on visual scene geometry (what is \textit{possible}).   \item \textbf{The Semantic Modulation:} The specific shift induced by the instruction.
\end{enumerate}

Standard VLA inference uses $s(a|o, l)$. However, we observe that $s(a|o, l) \approx s(a|o, \emptyset)$ when the language signal is weak or the visual features are overpowering.

To extract the \textbf{Pure Semantic Signal}, we calculate the \textbf{Residual Vector} $\Delta_{sem}$:
\begin{equation}
\Delta_{sem}(a, o, l) = s(a|o, l) - s(a|o, \emptyset)
\end{equation}

This subtraction cancels out the visual bias. If the robot wants to grasp a red cup ($a_{red}$) simply because it is close (visual bias), $s(a_{red}|o, \emptyset)$ will be high. The residual $\Delta_{sem}$ removes this bias, leaving only the text's contribution.

The final \textbf{Steered Policy} is:\begin{equation}\tilde{\pi}(a|o, l) \propto \exp\left( s(a|o, \emptyset) + \gamma \cdot \Delta_{sem}(a, o, l) \right)\end{equation}
where $\gamma > 1$ is the \textbf{Steering Coefficient}.

\textbf{Novelty vs. Standard Classifier-Free Guidance (CFG) \citep{ho2022classifier}:} While mechanically similar to CFG, RSS differs conceptually. CFG in diffusion acts as a quality booster (trading diversity for fidelity). RSS acts as a \textbf{Bias Suppressor}. We use the null-text pass to explicitly model the ``visual instinct'' of the robot and then mathematically penalize actions that are driven \textit{only} by that instinct and not confirmed by the text.

\subsection{Theoretical Analysis}

\textbf{Proposition 1 (Visual Bias Decoupling).} Let the logit function be approximated linearly as $S(o, l) = W_v \phi(o) + W_l \psi(l) + C$, where $C$ represents confounding interactions, $\phi(o) \in \mathbb{R}^d$ is the visual embedding vector, $\psi(l) \in \mathbb{R}^d$ is the linguistic embedding vector, $W_v, W_l \in \mathbb{R}^d$ are the projection weights for the visual and linguistic modalities, respectively, $\epsilon$ represents higher-order interaction terms and bias, assumed to be negligible for this first-order analysis. The standard inference yields a signal-to-noise ratio dependent on $\|W_v\| / \|W_l\|$. In VLA tasks, $\|W_v\| \gg \|W_l\|$.The Residual Steering yields:\begin{equation}\begin{split}\tilde{S} &= S(o, \emptyset) + \gamma(S(o, l) - S(o, \emptyset)) \\          &\approx W_v \phi(o) + \gamma W_l \psi(l)\end{split}\end{equation}

By setting $\gamma > 1$, we artificially restore the rank of the language features, effectively orthogonalizing the semantic vector from the visual manifold.

\section{Experiment} \label{sec:exp}

\begin{table*}[t]
\centering
\caption{
\textbf{Robustness evaluation under Destructive Instruction Overwriting.} 
Results are reported as task Success Rates (SR) in percentages (\%). 
\textbf{Bold} and \second{underlined} entries denote the best and second-best performance per column, respectively. 
Values in parentheses represent the absolute improvement over the corresponding baseline. 
\textbf{RAS}: Residual Affordance Steering; \textbf{MCSI}: Monte Carlo Syntactic Integration.
}\label{tab:libero_combine}
\resizebox{1.0\textwidth}{!}{
\begin{tabular}{c|ccccccccc|c}
\toprule
\multirow{2}{*}{\textbf{Model}} 
& \textbf{Origin}
& \textbf{Multi}
& \textbf{Blank}
& \textbf{Rand}
& \textbf{M2}
& \textbf{M4}
& \textbf{M6}
& \textbf{M8}
& \textbf{Simple}
& \textbf{Average} \\
& SR (\%) $\uparrow$ & SR (\%) $\uparrow$ & SR (\%) $\uparrow$ & SR (\%) $\uparrow$ & SR (\%) $\uparrow$ & SR (\%) $\uparrow$ & SR (\%) $\uparrow$ & SR (\%) $\uparrow$ & SR (\%) $\uparrow$ & SR (\%) $\uparrow$ \\
\midrule 
\multicolumn{11}{c}{$\pi_0$~\citep{black2024pi_0}} \\
\midrule
base                                     & 94.15 & 91.30 & 25.20 & 89.35 & 72.65 & 42.50 & 22.15 & 7.80  & 26.25 & 52.37 \\
\rowcolor{gray!20} + RAS                 & 90.65 & 89.90 & 63.40 & 88.70 & 78.45 & 55.05 & 33.55 & 17.95 & 62.50 & 64.46 \textcolor{red}{(+12.09)}\\
\rowcolor{gray!20} + MCSI             & 94.55 & 93.45 & 41.85 & 92.95 & 88.85 & 77.75 & 63.85 & 52.80 & 39.85 & 71.77 \textcolor{red}{(+19.40)}\\
\rowcolor{gray!20} + RAS \& MCSI      & 93.35 & 91.35 & 69.65 & 94.95 & 92.05 & 85.85 & 77.95 & 69.90 & 64.90 & 82.22 \textcolor{red}{(+29.85)}\\
\midrule
\multicolumn{11}{c}{$\pi_{0.5}$~\citep{intelligence2025pi05}} \\
\midrule
base                                     & 95.15 & 95.45 & 50.05 & 95.20 & 92.65 & 82.70 & 69.55 & 55.45 & 46.90 & 75.90 \\
\rowcolor{gray!20} + RAS                 & \second{96.65} & 96.80 & \textbf{70.50} & \second{95.95} & 93.90 & 86.65 & \second{79.30} & \second{71.05} & \second{69.10} & \second{84.43} \textcolor{red}{(+8.53)} \\
\rowcolor{gray!20} + MCSI             & \textbf{98.25} & \textbf{98.00} & 46.20 & \textbf{97.45} & \second{96.05} & \second{87.20} & 73.40 & 57.95 & 46.20 & 77.86 \textcolor{red}{(+1.96)}\\
\rowcolor{gray!20} + RAS \& MCSI      & 96.60 & \second{97.50} & \second{70.25} & \textbf{97.45} & \textbf{96.35} & \textbf{92.00} & \textbf{84.55} & \textbf{77.50} & \textbf{70.60} & \textbf{86.98} \textcolor{red}{(+11.08)}\\
\bottomrule
\end{tabular}}
\end{table*}

\begin{table*}
\centering
\caption{
\textbf{Obfuscated instruction reinterpretation result.} \textbf{Bold} entries denote the best results per column, and \second{underlined entries} indicate the second-best. Values in parentheses report the absolute SR improvement (in percentage points) over the corresponding baseline under the same setting. All numbers are percentages (\%). \textbf{RAS}: Residual Affordance Steering; \textbf{MCSI}: Monte Carlo Syntactic Integration.
}
\label{tab:libero_goal_variant}

\resizebox{0.75\textwidth}{!}{
\begin{tabular}{c|ccccc|c}
\toprule
\multirow{2}{*}{\textbf{Model}} & \textbf{R0} & \textbf{R1} & \textbf{R2} & \textbf{R3} & \textbf{R4} & \textbf{Average} \\
& SR (\%) $\uparrow$ & SR (\%) $\uparrow$ & SR (\%) $\uparrow$ & SR (\%) $\uparrow$ & SR (\%) $\uparrow$ & SR (\%) $\uparrow$ \\

\midrule 
\multicolumn{7}{c}{$\pi_0$~\citep{black2024pi_0}} \\
\midrule
base                             & 91.4 & 55.8 & 7.4 & 28.4 & 42.4 & 45.08 \\
\rowcolor{gray!20}  + RAS                          & 90.0 & 59.0 & 11.4 & 40.2 & 46.0 & 49.32 \textcolor{red}{(+4.24)} \\
\rowcolor{gray!20}  + MCSI                      & 92.6 & 83.6 & 28.0 & 71.2 & 58.6 & 66.80 \textcolor{red}{(+21.72)}\\
\rowcolor{gray!20}  + RAS \& MCSI               & 88.4 & 85.4 & 26.8 & 80.0 & 47.0 & 65.52 \textcolor{red}{(+20.44)}\\

\midrule
\multicolumn{7}{c}{$\pi_{0.5}$~\citep{intelligence2025pi05}} \\
\midrule
base                 & 95.0 & 93.2 & \second{30.4} & \textbf{90.6} & 68.6 & 75.56 \\
\rowcolor{gray!20}  + RAS                      & \second{97.0} & \second{94.6} & 26.4 & 86.4 & \second{77.6} & \second{76.40} \textcolor{red}{(+0.84)}\\
\rowcolor{gray!20}  + MCSI                  & \second{97.0} & \second{94.6} & \textbf{31.4} & 84.4 & 70.6 & 75.60 \textcolor{red}{(+0.04)} \\
\rowcolor{gray!20}  + RAS \& MCSI           & \textbf{97.6} & \textbf{97.2} & 30.2 & \second{89.4} & \textbf{79.0} & \textbf{78.68} \textcolor{red}{(+3.12)} \\
\bottomrule
\end{tabular}}
\end{table*}

\begin{table*}
\centering
\caption{ \textbf{Examples of obfuscated instruction reinterpretation variants (R0–R4) on LIBERO-Goal.} All variants preserve the original task semantics while introducing different forms of linguistic perturbations.}
\label{tab:instruction_rewriting}
\resizebox{1.0\textwidth}{!}{
\begin{tabular}{l l p{0.75\linewidth}}
\toprule
\textbf{Variant}  & \textbf{Function} & \textbf{Examples} \\
\midrule
Original & / & Put the wine bottle on top of the cabinet. \\
R0 & Multiword Substitution & Move the wine bottle onto the cabinet top. \\
R1 & Distraction & Once you've got a steady hold, put the wine bottle on the cabinet top. \\
R2 & Common Sense & Set the sealed container typically used for pouring grape-based drinks on the upper face of the standing storage furniture. \\
R3 & Reasoning Chain & Move the bottle over the cabinet, then release it once it is stable on top. \\
R4 & Confusion & Regardless of the drawers being open or closed, place the wine bottle on top of the cabinet. \\
\bottomrule
\end{tabular}
}
\end{table*}

\begin{table}
\centering
\caption{
\textbf{Out-of-distribution semantic transfer result.} \textbf{Bold} entries denote the best results per column. The base model is $\pi_{0.5}$~\citep{intelligence2025pi05}. \textbf{RAS}: Residual Affordance Steering; \textbf{MCSI}: Monte Carlo Syntactic Integration.
}
\label{tab:libero_goal_ood_fewstep}
\resizebox{0.5\textwidth}{!}{
\begin{tabular}{c|ccc|c}
\toprule
\multirow{2}{*}{\textbf{Model}} 
& \textbf{10-steps} 
& \textbf{100-steps} 
& \textbf{1000-steps} 
& \textbf{Average} \\
& SR (\%) $\uparrow$ 
& SR (\%) $\uparrow$ 
& SR (\%) $\uparrow$ 
& SR (\%) $\uparrow$ \\
\midrule

base                   & 27.0 & 31.0 & 91.0 &  49.67  \\
\rowcolor{gray!20}  + RAS                        & 17.0 & 29.0 & \textbf{98.0} &   48.00 \\
\rowcolor{gray!20}  + MCSI                    & 28.0 & \textbf{42.0} & 97.0 &   \textbf{55.67} \\
\rowcolor{gray!20} + RAS \& MCSI             & \textbf{31.0} & 31.0 & 97.0 &  53.00  \\

\bottomrule
\end{tabular}}
\end{table}

\begin{figure}[t]
  \centering
  \includegraphics[width=0.5\textwidth]{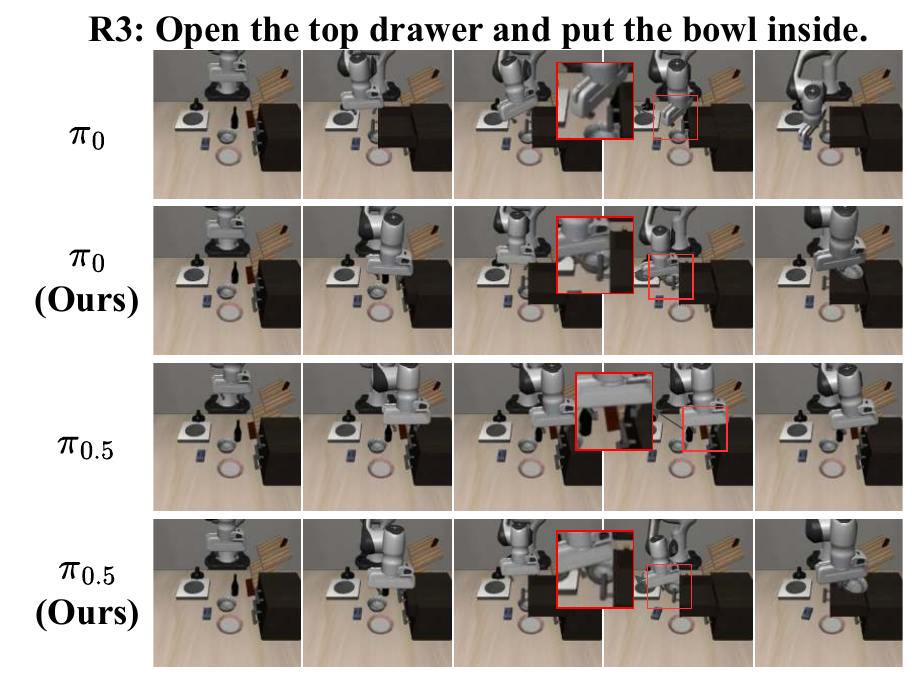}
  \caption{
   \textbf{Comparison on the LIBERO variant \textit{R3-Reasoning Chain}.}
In the "open the top drawer and put the bowl inside" task, our model consistently outperforms the baseline under reasoning-chain–perturbed instructions, demonstrating a stronger ability to follow multi-step semantic constraints and accurately complete the task despite increased linguistic complexity.
  }
  \label{fig:varient_r3_compare}
\end{figure}

\subsection{Training Setting}
We adopt $\pi_0$~\citep{black2024pi_0} and $\pi_{0.5}$~\citep{intelligence2025pi05} as our baseline models. The pretrained vision language model backbone is based on Gemma~\citep{team2024gemma}. The initial base weights are obtained through large-scale pretraining on a diverse and heterogeneous collection of robotic datasets~\cite{o2024openx}. Building upon these pretrained representations, we fine-tune the models on task-specific data and incorporate our proposed modifications.

To implement \emph{Monte Carlo Syntactic Integration}, we utilize Qwen2.5-VL~\cite{bai2025qwen25vl} as an oracle teacher to generate dense semantic neighborhoods during training. At evaluation time, we employ ChatGPT-5.2 \citep{openai_chatgpt} to systematically rewrite all task instructions, ensuring consistent and controlled linguistic variations across experiments.

Each model was trained for a total of $30{,}000$ steps with a batch size of $32$. The learning rate followed a cosine decay schedule, implemented as \emph{CosineDecaySchedule}, with a warm-up phase of $10{,}000$ steps, a peak learning rate of $5\times10^{-5}$ and a final learning rate of $5\times10^{-5}$. An exponential moving average (EMA) with a decay rate of $0.999$ was applied. During inference, a single NVIDIA RTX~3090 GPU was used for model evaluation and deployment on the server side.

\subsection{Simulation Experiment}

We adopt \textbf{LIBERO}~\cite{liu2023libero} as our primary simulation benchmark for evaluating VLA models. It comprises four task categories (\textit{LIBERO-Spatial}, \textit{LIBERO-Object}, \textit{LIBERO-Goal}, and \textit{LIBERO-10}) which jointly challenge VLA models along multiple dimensions, including spatial reasoning, object-centric manipulation, goal specification, and multi-step task execution. As a result, it has become a highly competitive benchmark, with many state-of-the-art methods employing sophisticated architectural designs and advanced training strategies to maximize performance.

To further investigate how natural language instructions guide VLA models, we extend LIBERO with a set of \emph{controlled instruction variants} that explicitly probe linguistic robustness and generalization. Specifically, we consider three categories of instruction perturbations (See Figure~\ref{fig:perturb_examples}).
(1) \textbf{Destructive instruction overwriting}, where critical semantic components of the original instruction are intentionally corrupted or removed, thereby disrupting the instruction's compositional meaning. 
(2) \textbf{Obfuscated instruction reinterpretation}, which introduces syntactically valid yet semantically peripheral or potentially distracting descriptions while preserving the original task intent to evaluate robustness against semantic distraction rather than outright corruption. 
(3) \textbf{Out-of-distribution (OOD) semantic transfer}, where novel task instructions are constructed by recombining object concepts observed during training into previously unseen compositions, resulting in tasks that do not appear in the original training distribution.

\paragraph{Destructive instruction overwriting.}
To explicitly probe the extent to which VLA policies rely on genuine language grounding rather than spurious correlations, we design a set of \emph{destructive instruction overwriting} variants that deliberately corrupt or erase critical semantic information in the original task instructions.
Specifically, we consider five types of perturbations:
\textbf{Blank}, where the instruction is replaced by an empty string, completely removing linguistic input;
\textbf{Simple}, where all instructions are substituted with a generic and uninformative phrase (e.g., ``Do something'');
\textbf{Multi}, where we leverage a large language model to generate multiple paraphrases of the original instruction via word substitution or rephrasing, and randomly sample one variant during evaluation;
\textbf{Rand}, which randomly permutes the word order within the instruction, disrupting syntactic structure while preserving the vocabulary;
and \textbf{Mask}, where each word is independently replaced by a \texttt{<MASK>} token with varying probabilities, progressively degrading semantic content.

\begin{figure*}[t]
  \centering
  \includegraphics[width=1.\textwidth, height=0.8\textheight, keepaspectratio]{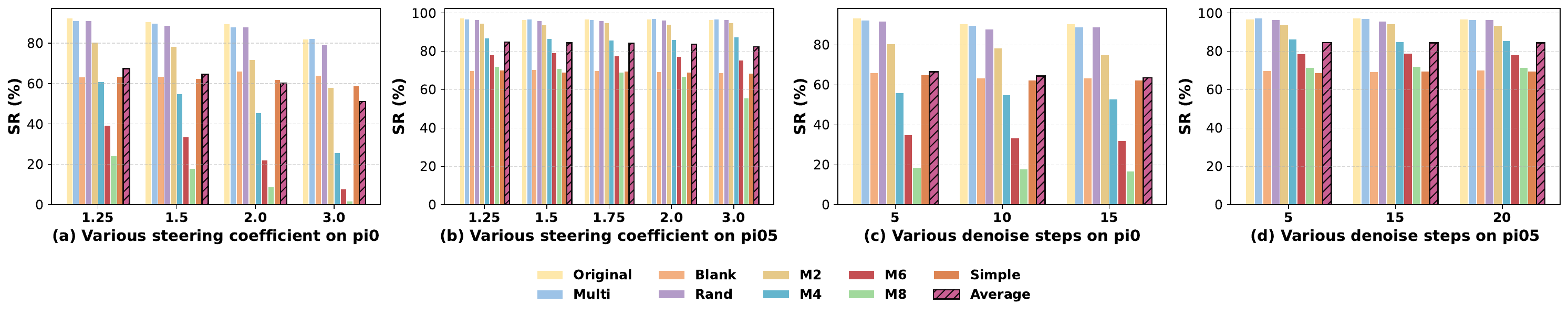}
  \caption{
  \textbf{Ablation of steering coefficient and denoising steps on destructive instruction overwriting.}
Success rates (SR, \%) across instruction variants under different steering coefficients for $\pi_0$ (a) and $\pi_{0.5}$ (b), and different denoising steps for $\pi_0$ (c) and $\pi_{0.5}$ (d), illustrating the effect of guidance and generation depth on robustness to instruction perturbations.
  }
  \label{fig:ablation}
\end{figure*}

As shown in Table~\ref{tab:libero_combine}, these destructive perturbations lead to substantial performance degradation across most settings, particularly under \textbf{Blank}, \textbf{Simple}, and high-ratio \textbf{Mask} conditions, indicating that removing or severely corrupting linguistic cues significantly hinders task execution. Notably, models augmented with RAS and MCSI demonstrate improved resilience under destructive overwriting, achieving higher average success rates across all variants. This observation implies that richer vision–language alignment can partially mitigate the reliance on brittle instruction patterns, encouraging more semantically grounded policy behavior rather than shortcut exploitation.

\paragraph{Obfuscated instruction reinterpretation.}
We introduce a set of semantically preserved but potentially obfuscated instruction variants to evaluate linguistic robustness on the LIBERO-Goal benchmark.
Specifically, we consider:
\textbf{R0 (Multiword Substitution)}, which applies lightweight synonym or phrase replacements;
\textbf{R1 (Distraction)}, which augments instructions with task-irrelevant contextual content;
\textbf{R2 (Common Sense)}, which replaces object names with commonsense-based descriptive phrases;
\textbf{R3 (Reasoning Chain)}, which reformulates instructions to emphasize implicit reasoning or final-state constraints;
and \textbf{R4 (Confusion)}, which introduces distractor objects via explicit negation while preserving the original task goal.

As shown in Table~\ref{tab:libero_goal_variant}, there are clear performance disparities across different reinterpretation variants, highlighting how VLA models respond to semantically preserved yet linguistically challenging instructions.
Among all settings, \textbf{R0} incurs only minor degradation, indicating that most models are largely insensitive to superficial lexical substitutions.
In contrast, \textbf{R1} and \textbf{R2} lead to more pronounced drops, suggesting that additional irrelevant context and abstract commonsense descriptions substantially increase the difficulty of extracting task-relevant semantics.

Notably, \textbf{R3} and \textbf{R4} constitute the most challenging conditions.
\textbf{R3} requires the model to correctly interpret implicit reasoning structures and final-state constraints, while \textbf{R4} explicitly introduces distractor objects that co-occur in other tasks, stressing the model’s ability to resist spurious correlations.
Performance under these two variants, therefore, serves as a stronger indicator of true language grounding rather than shallow pattern matching.

Across most variants, models augmented with RAS and MCSI demonstrate improved robustness, achieving the highest average success rate. This trend suggests that richer vision–language alignment encourages policies to rely less on brittle lexical cues and more on semantically grounded representations, improving generalization under obfuscated yet valid instructions.

\begin{figure}[t]
  \centering
  \includegraphics[width=0.5\textwidth]{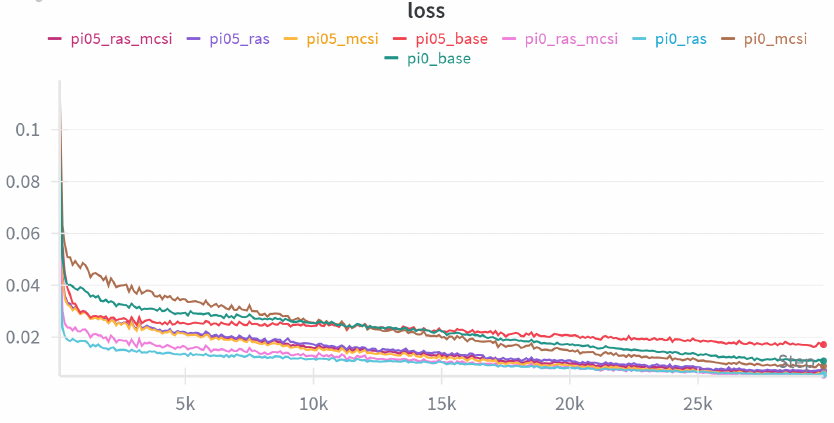}
  \caption{
  \textbf{Training loss curves.}
  We report the training loss trajectories of different model variants throughout optimization. 
  \textbf{RAS}: Residual Affordance Steering; \textbf{MCSI}: Monte Carlo Syntactic Integration.
  }
  \label{fig-train_loss_curve}
\end{figure}

\paragraph{Out-of-distribution semantic transfer.}

To construct a controlled out-of-distribution (OOD) evaluation setting, we remove two target tasks from the original LIBERO-Goal training set, while ensuring that the objects involved still appear in other training tasks. This design isolates distribution shift at the level of \emph{object–goal composition}, rather than introducing entirely unseen objects. To avoid trivial failure due to complete task absence, we first fine-tune the baseline model for 6{,}000 steps using the remaining in-distribution data, corresponding to approximately 20\% of full training. We then perform few-step adaptation using a small number of demonstrations from the two held-out OOD tasks, and evaluate performance under varying step budgets.

As shown in Table~\ref{tab:libero_goal_ood_fewstep}, the vanilla model exhibits limited generalization under low-step regimes, particularly with 10-steps adaptation, indicating difficulty in transferring previously learned object semantics to novel task compositions. Meanwhile, although the average success rate of the baseline appears relatively high (in 100-steps and 1,000-steps adaptation), a closer inspection reveals that this performance is dominated by overfitting to a single held-out task, while consistently failing on the other OOD task. In contrast, our method achieves successful execution across both held-out tasks, reflecting improved robustness to compositional distribution shifts.

Notably, RAS and MCSI improve few-step performance to different extents across regimes, with MCSI consistently achieving stronger gains. While combining RAS with MCSI offers additional benefits, MCSI alone delivers the most effective improvement, indicating its dominant role in enhancing semantic transfer and reducing reliance on task-specific memorization.

\subsection{Ablation Study}

We conduct a comprehensive ablation study on two key hyperparameters in RSS: the Residual Affordance Steering (RAS) coefficient and the number of denoising steps. All evaluations are performed under destructive instruction overwriting benchmarks.

\textbf{Residual Affordance Steering.}
Figure~\ref{fig:ablation} (a) and Figure~\ref{fig:ablation} (b) report performance under different RAS coefficients with the number of denoising steps fixed to 10. We observe that moderate steering coefficients consistently improve robustness under semantic-preserving perturbations, by strengthening the alignment between language conditions and action generation. However, excessively large coefficients significantly amplify sensitivity to corrupted or obfuscated instructions, leading to pronounced performance drops under destructive perturbations. This trend indicates that over-strong RAS encourages over-conditioning on unreliable linguistic signals, thereby exacerbating shortcut exploitation. More details can be found in the Table~\ref{tab:ablation_sc}.

\textbf{Denoising steps.}
Figure~\ref{fig:ablation} (c) and Figure~\ref{fig:ablation} (d) analyze the effect of varying the number of denoising steps with the RAS coefficient fixed at 1.5 and 1.25, respectively. While increasing the number of steps can improve success rates in certain individual settings, we observe that larger step counts do not consistently translate to higher overall performance, and in some cases lead to a slight degradation in average success rate. This suggests that excessive denoising primarily refines low-level action details without providing additional semantic benefits.

\textbf{Joint effect.}
When considering the interaction between RAS and denoising steps, we find that the overall impact of varying the number of denoising steps is relatively limited under smaller steering coefficients. In this regime, the model is less sensitive to sampling depth, as weaker conditional forcing mitigates over-reliance on potentially noisy language inputs. Overall, the optimal configuration reflects a trade-off between conditional strength and sampling stability, with moderate RAS coefficients and a moderate number of denoising steps yielding the most robust performance across language perturbations.

\subsection{Different VLM Teachers.}

To demonstrate that MCSI is not merely overfitting to the specific linguistic patterns of the training teacher, we conducted the following analyses without retraining (See Table~\ref{tab:libero_diff_vlm_eval}).

\begin{table*}[htbp]
\centering
\caption{
\textbf{Generalization to paraphrased instructions generated by different VLMs on LIBERO. }
Without retraining, \textbf{MCSI} consistently improves robustness to linguistic variations 
and reduces performance drift compared to the baseline $\pi_{0.5}$~\citep{intelligence2025pi05}. 
\textbf{RAS}: Residual Affordance Steering; \textbf{MCSI}: Monte Carlo Syntactic Integration.
}
\label{tab:libero_diff_vlm_eval}

\resizebox{0.68\textwidth}{!}{
\begin{tabular}{c|cccc|cc}
\toprule
\multirow{2}{*}{\textbf{Model}} & \textbf{R1} & \textbf{R2} & \textbf{R3} & \textbf{R4} & \textbf{Average} & \multirow{2}{*}{\textbf{Drift}} \\
& SR (\%) $\uparrow$ & SR (\%) $\uparrow$ & SR (\%) $\uparrow$ & SR (\%) $\uparrow$ & SR (\%) $\uparrow$ & \\

\midrule 
\multicolumn{7}{c}{Instructions generated by \textbf{ChatGPT5.2~\cite{openai_chatgpt}}} \\
\midrule
 base                                           & 93.2 & 30.4 & 90.6 & 68.6 & 70.70 & 0 \\
\rowcolor{gray!20}  + MCSI                      & 94.6 & 31.4 & 84.4 & 70.6 & 70.25 & 0 \\
\rowcolor{gray!20}  + RAS \& MCSI               & 97.2 & 30.2 & 89.4 & 79.0 & 73.95 & 0 \\

\midrule
\multicolumn{7}{c}{Instructions generated by \textbf{DeepSeek-R1~\cite{guo2025deepseek}}} \\
\midrule
base                                         & 84.8 & 26.6 & 84.4 & 57.2 & 63.25 & \textcolor{blue}{-10.54\%} \\
\rowcolor{gray!20}  + MCSI                   & 92.2 & 35.0 & 81.8 & 68.4 & 69.35 & \textcolor{blue}{-1.28\%}  \\
\rowcolor{gray!20}  + RAS \& MCSI            & 92.2 & 26.6 & 87.2 & 77.4 & 70.85 & \textcolor{blue}{-4.19\%}  \\

\midrule
\multicolumn{7}{c}{Instructions generated by \textbf{Qwen3.5~\cite{qwen3.5}}} \\
\midrule
base                                         & 90.0 & 32.8 & 72.8 & 72.8 & 67.10 & \textcolor{blue}{-5.10\%}  \\
\rowcolor{gray!20}  + MCSI                   & 95.0 & 34.4 & 71.6 & 75.2 & 69.05 & \textcolor{blue}{-1.71\%}  \\
\rowcolor{gray!20}  + RAS \& MCSI            & 95.0 & 34.4 & 85.4 & 85.2 & 75.00 & \textcolor{red}{+1.41\%}  \\

\bottomrule
\end{tabular}}
\end{table*}

\begin{table*}[h]
\centering
\caption{
\textbf{Utilize different VLMs for instruction generation to train the MCSI \& RAS model.}
\textbf{MCSI} and \textbf{RAS} shows consistent performance across VLM choices with minimal variation, 
indicating robustness to the choice of instruction generator and generalization to underlying task semantics. 
}

\label{tab:libero_diff_vlm_train}

\resizebox{0.8\textwidth}{!}{
\begin{tabular}{c|cccc|c|c}
\toprule
\multirow{2}{*}{\textbf{VLM}} & \textbf{R1} & \textbf{R2} & \textbf{R3} & \textbf{R4} & \textbf{Average} & \multirow{2}{*}{\textbf{Drift}} \\
& SR (\%) $\uparrow$ & SR (\%) $\uparrow$ & SR (\%) $\uparrow$ & SR (\%) $\uparrow$ & SR (\%) $\uparrow$ & \\
\midrule
Qwen2.5-VL~\cite{bai2025qwen25vl}                     & 97.2 & 30.2 & 89.4 & 79.0 & 73.95 & 0 \\
InternVL3~\cite{chen2024internvl}                  & 96.8 & 34.0 & 93.0 & 79.2 & 75.75 & \textcolor{red}{+2.43\%} \\
LLaVA-OneVision~\cite{lillava}                & 96.2 & 33.6 & 91.2 & 73.2 & 73.55 & \textcolor{blue}{-0.54\%} \\

\bottomrule
\end{tabular}}
\end{table*}

We tested the MCSI model on instructions paraphrased by DeepSeek-R1~\cite{guo2025deepseek} and Qwen-3.5~\cite{qwen3.5}. Despite distinct lexical choices and sentence structures, MCSI demonstrated significantly higher stability than the baseline. Specifically, when instructions were rephrased by DeepSeek-R1, the baseline $\pi_{0.5}$ model suffered a performance drop of 10.54\%, whereas our method (add \text{MCSI}) declined by only 1.28\%. Similarly, with Qwen-3.5, the baseline dropped 5.10\% compared to just 1.71\% for our method.

Beyond using VLMs for rewriting, we also investigate using different VLMs for instruction generation: Beyond the original Qwen2.5-VL~\cite{bai2025qwen25vl}, we tested InternVL3~\cite{chen2024internvl} and LLaVA-OneVision~\cite{lillava} for generating syntactic neighborhoods. The performance variance was minimal (2.43\% and 0.54\%, respectively), confirming that current VLMs provide similarly effective spatial understanding for our method (See Table~\ref{tab:libero_diff_vlm_train}).

These results empirically confirm that MCSI generalizes to the underlying semantic intent of the task rather than overfitting to the specific syntax of the training teacher.

\subsection{Training Loss Curves}

Figure~\ref{fig-train_loss_curve} illustrates the training loss trajectories of different model variants. Across all settings, the loss decreases rapidly during the early training stage and gradually stabilizes as optimization proceeds, indicating stable convergence behavior. Compared to their corresponding baseline models, variants equipped with RAS and MCSI achieve lower loss values when training. This suggests that the proposed components provide a more effective training signal and facilitate smoother optimization, leading to improved learning efficiency rather than merely affecting the final performance.

\section{Conclusion} \label{sec:conclusion}

We identify a core limitation of current VLA models: their failure to robustly ground linguistic intent, resulting in instruction blindness and overreliance on visual affordance priors. To address this, we propose \textbf{Residual Semantic Steering (RSS)}, which combines \textbf{Monte Carlo Syntactic Integration} to overcome instruction manifold sparsity with \textbf{Residual Affordance Steering} to suppress visually driven biases. By explicitly disentangling semantic intent from visual priors, RSS restores consistent language–action alignment and enables robust generalization under diverse linguistic perturbations. Empirical results demonstrate that RSS significantly improves instruction robustness and semantic grounding, establishing a principled path toward reliable language-conditioned robotic control.

\clearpage
\section*{Limitations} \label{sec:limitation}

A limitation of Residual Affordance Steering (RAS) is its conservative behavior when presented with extremely vague or underspecified instructions (e.g., generic prompts like ``do something''). Because our inference mechanism explicitly suppresses the ``Base Affordance Distribution''—which drives standard models to aggressively execute the most common visual trajectory regardless of input—our model may exhibit hesitation or inaction when the language signal lacks sufficient semantic content to steer the policy. Unlike baseline methods that tend to ignore linguistic ambiguity and revert to rote visual pattern matching (often executing an action simply because it was frequent in the training set), our framework effectively requires semantically meaningful commands to initiate motion. This dependency prevents the model from ``hallucinating'' actions based on visual priors alone, ensuring that the robot does not default to unsafe autopilot behaviors when the user's intent is unclear.

\section*{Acknowledgments}

This work was supported in part by National High-Level Young Talent Program (Grant 2025HY00260104), in part by the Fundamental Research Funds for Higher Education Institutions allocated to Sun Yat-sen University (Grant 25hytd007), in part by the Guangdong Provincial High-Level Young Talent Program (Grant 2025HYSPT0707), in part by the Tuoyuan (Grant HT-99982025-0564), in part by the Faculty Start-up Research Fund (Grant 67000-12255002), in part by the Huawei Strategic Research Institute Talent Fund, and in part by the Key Development Project of the Artificial Intelligence Institute of Sun Yat-sen University (Grant 2025RGZN009).

\bibliography{custom}

\appendix

\section{Proof of Proposition 1: Visual Bias Decoupling via Residual Steering}
\label{appendix:proof}

In this section, we provide a formal derivation of Proposition 1, demonstrating how Residual Semantic Steering (RSS) restores the rank of linguistic features in the presence of dominant visual priors.

\subsection{Setup and Definitions}

Let $S(a | o, l) \in \mathbb{R}$ denote the logit score for a specific action $a$, given observation $o$ and instruction $l$. We analyze the behavior of the model's final layer. We assume the logit can be approximated as a linear combination of the disentangled features in the penultimate layer:

\begin{equation}
    S(a | o, l) = W_v^\top \phi(o) + W_l^\top \psi(l) + \epsilon
\end{equation}

where:
\begin{itemize}
    \item $\phi(o) \in \mathbb{R}^d$ is the visual embedding vector.
    \item $\psi(l) \in \mathbb{R}^d$ is the linguistic embedding vector.
    \item $W_v, W_l \in \mathbb{R}^d$ are the projection weights for the visual and linguistic modalities, respectively.
    \item $\epsilon$ represents higher-order interaction terms and bias, assumed to be negligible for this first-order analysis.
\end{itemize}

\textbf{Assumption 1 (Visual Dominance):} In VLA models, the gradient flow during training is dominated by the dense visual signal, leading to spectral norms such that $\|W_v\| \gg \|W_l\|$. This implies that standard inference is driven primarily by $\phi(o)$.

\textbf{Assumption 2 (Null-Text Baseline):} When the instruction is dropped (denoted as $\emptyset$), the linguistic feature vector collapses to a null or bias state, denoted as $\psi(\emptyset) \approx \mathbf{0}$. Thus, the unconditional forward pass represents the pure visual affordance:
\begin{equation}
    S(a | o, \emptyset) \approx W_v^\top \phi(o)
\end{equation}

\subsection{Derivation of Residual Steering}

The Residual Semantic Steering formula is defined as:
\begin{equation}
    \tilde{S}(a) = S(a | o, \emptyset) + \gamma \left( S(a | o, l) - S(a | o, \emptyset) \right)
\end{equation}
where $\gamma$ is the steering coefficient.

Substituting the linear approximations into the residual term:
\begin{align}
    \Delta_{sem} &= S(a | o, l) - S(a | o, \emptyset) \\
    &= (W_v^\top \phi(o) + W_l^\top \psi(l)) \\ 
    & \ \ \ \ \ \ - (W_v^\top \phi(o) + W_l^\top \psi(\emptyset)) \\
    &= W_l^\top (\psi(l) - \mathbf{0}) \\
    &= W_l^\top \psi(l)
\end{align}
Note that the visual term $W_v^\top \phi(o)$ is explicitly cancelled out in the residual term.

Now, we reconstruct the full steered logit $\tilde{S}(a)$:
\begin{align}
    \tilde{S}(a) &= S(a | o, \emptyset) + \gamma \cdot \Delta_{sem} \\
    &= W_v^\top \phi(o) + \gamma (W_l^\top \psi(l))
\end{align}

\subsection{Analysis of Signal-to-Noise Ratio (SNR)}

We define the Semantic Signal-to-Noise Ratio (SNR) as the ratio of the linguistic contribution to the visual contribution.

\textbf{Case 1: Standard Inference ($\gamma = 1$)}
\begin{equation}
    \text{SNR}_{std} = \frac{| W_l^\top \psi(l) |}{| W_v^\top \phi(o) |}
\end{equation}
Given Assumption 1 ($\|W_v\| \gg \|W_l\|$), $\text{SNR}_{std} \to 0$. The text has minimal impact on the action ranking.

\textbf{Case 2: Residual Steering ($\gamma > 1$)}
\begin{equation}
    \text{SNR}_{rss} = \frac{| \gamma W_l^\top \psi(l) |}{| W_v^\top \phi(o) |} = \gamma \cdot \text{SNR}_{std}
\end{equation}

\subsection{Conclusion}
By choosing $\gamma \gg 1$, we linearly amplify the linguistic contribution without altering the visual affordance landscape. Effectively, we synthesize a new weight vector $\tilde{W}_l = \gamma W_l$, artificially restoring the balance between modalities. This proves that RSS orthogonalizes the semantic intent from the visual prior, as stated in Proposition 1. \hfill $\square$

\section{More results on LIBERO-Plus} \label{appendix:libero_plus_result}

To more convincingly demonstrate the effectiveness of the proposed method, we evaluate our model on LIBERO-Plus~\cite{fei2025libero} using the trained checkpoint without any additional retraining.

As shown in the Table~\ref{tab:libero_plus_results}, our method demonstrates superior performance. Crucially, it not only outperforms baselines in language-perturbed subsets but also achieves the best overall performance across the benchmark compared to competitors. This confirms that our semantic integration method improves robustness without becoming sensitive to the visual domain shifts present in Libero-Plus. We will include these results in the final revision.

\begin{table*}[htbp]
\centering
\caption{
\textbf{LIBERO-Plus task performance (success rate, SR) results.} \textbf{Bold} denotes the best performance per column, respectively. 
Values in parentheses represent the absolute improvement over the corresponding baseline. \textbf{RAS}: Residual Affordance Steering; \textbf{MCSI}: Monte Carlo Syntactic Integration.
}
\label{tab:libero_plus_results}
\resizebox{0.90\textwidth}{!}{
\begin{tabular}{c|ccccccc|c}
\toprule
\multirow{2}{*}{\textbf{Model}} & \textbf{Camera} & \textbf{Robot} & \textbf{Language} & \textbf{Light} & \textbf{Background} & \textbf{Noise} & \textbf{Layout} & \textbf{Average} \\
& SR (\%) $\uparrow$ & SR (\%) $\uparrow$ & SR (\%) $\uparrow$ & SR (\%) $\uparrow$ & SR (\%) $\uparrow$ & SR (\%) $\uparrow$ & SR (\%) $\uparrow$ & SR (\%) $\uparrow$\\
\midrule
OpenVLA~\cite{kim2024openvla}              & 0.8  & 3.5  & 23.0 & 8.1  & 34.8 & 15.2 & 28.5 & 15.6 \\
WorldVLA~\cite{cen2025worldvla}            & 0.1  & 27.9 & 41.6 & 43.7 & 17.1 & 10.9 & 38.0 & 25.0 \\
NORA~\cite{hung2025nora}                   & 2.2  & 37.0 & 65.1 & 45.7 & 58.6 & 12.8 & 62.1 & 39.0 \\
UniVLA~\cite{bu2025univla}                 & 1.8  & 46.2 & 69.6 & 69.0 & 81.0 & 21.2 & 31.9 & 43.9 \\
$\pi_0$~\cite{black2024pi_0}               & 13.8 & 6.0  & 58.8 & 85.0 & 81.4 & 79.0 & 68.9 & 53.6 \\
OpenVLA-OFT (w)~\cite{kim2025fine}         & 10.4 & 38.7 & 70.5 & 76.8 & 93.6 & 49.9 & 69.9 & 55.8 \\
$\pi_0$ FAST \cite{pertsch2025pi0fast}        & 65.1 & 21.6 & 61.0 & 73.2 & 73.2 & 74.4 & 68.8 & 61.6 \\
OpenVLA-OFT (m)~\cite{kim2025fine}         & 55.6 & 21.7 & 81.0 & 92.7 & 91.0 & 78.6 & 68.7 & 67.9 \\
OpenVLA-OFT~\cite{kim2025fine}             & 56.4 & 31.9 & 79.5 & 88.7 & 93.3 & 75.8 & 74.2 & 69.6 \\
RIPT-VLA~\cite{tan2025interactive}         & 55.2 & 31.2 & 77.6 & 88.4 & 91.6 & 73.5 & 74.2 & 68.4 \\
OpenVLA-OFT (plus)~\cite{fei2025libero}& \textbf{92.8} & 30.3 & 85.8 & 94.9 & 93.9 & 89.3 & 77.6 & 79.6 \\
\midrule
$\pi_{0.5}$~\cite{intelligence2025pi05}    & 64.8 & 71.8 & 83.0 & 93.5 & 92.2 & 78.8 & 85.5 & 81.4 \\
\rowcolor{gray!20}
 $\pi_{0.5}$ + RAS                & 76.0 & 74.0 & 88.0 & \textbf{97.0} & 96.0 & 83.0 & 86.0 & 86.0 \textcolor{red}{(+4.6)} \\
\rowcolor{gray!20}
 $\pi_{0.5}$ + MCSI                & 86.0 & 83.0 & 83.0 & \textbf{97.0} & \textbf{98.0} & \textbf{92.0} & \textbf{87.0} & \textbf{90.0} \textcolor{red}{(+8.6)} \\
\rowcolor{gray!20}
 $\pi_{0.5}$ + RAS \& MCSI             & 81.0 & \textbf{88.0} & \textbf{90.0} & \textbf{97.0} & 96.0 & 90.0 & 86.0 & \textbf{90.0} \textcolor{red}{(+8.6)} \\
\bottomrule
\end{tabular}}
\end{table*}

\section{Detailed Results} \label{appendix:detailed_result}

This section provides a comprehensive breakdown of task-level performance across all LIBERO subtasks. Unlike the aggregated results reported in the main paper, the tables in this section present success rates for each individual subtask, enabling a fine-grained examination of model behavior under different instruction conditions.

Specifically, Table~\ref{tab:libero_origin} reports performance under the original, unmodified instructions, serving as a reference for canonical task execution. Tables~\ref{tab:libero_blank}, \ref{tab:libero_simple}, \ref{tab:libero_multi}, \ref{tab:libero_random}, and~\ref{tab:libero_mask} report detailed subtask-level success rates under different instruction perturbation settings.

\begin{table*}[htbp]
\centering
\caption{
\textbf{LIBERO (original instruction) task performance (success rate, SR) results.} Results are reported on the four standard LIBERO sub-tasks (Libero-Spatial, Libero-Object, Libero-Goal, and Libero-Long) along with the overall average. This setting evaluates model performance under unmodified, canonical task instructions.
\textbf{RAS}: Residual Affordance Steering; \textbf{MCSI}: Monte Carlo Syntactic Integration.
}
\label{tab:libero_origin}
\resizebox{0.8\textwidth}{!}{
\begin{tabular}{c|cccc|c}
\toprule
\multirow{2}{*}{\textbf{Model}} & \textbf{Libero-Spatial} & \textbf{Libero-Object} & \textbf{Libero-Goal} & \textbf{Libero-Long} & \textbf{Average} \\
& SR (\%) $\uparrow$ & SR (\%) $\uparrow$ & SR (\%) $\uparrow$ & SR (\%) $\uparrow$ & SR (\%) $\uparrow$ \\
\midrule
Diffusion Policy~\citep{chi2023diffusion}                 & 78.3 & 92.5 & 68.3 & 50.5 & 72.40 \\
MDT~\citep{reuss2024multimodal}                           & 78.5 & 87.5 & 73.5 & 64.8 & 76.10 \\
OpenVLA~\citep{kim2024openvla}                            & 84.7 & 88.4 & 79.2 & 53.7 & 76.50 \\
Octo \citep{mees2024octo}                                 & 78.9 & 85.7 & 84.6 & 51.1 & 75.10 \\
Dita~\citep{hou2025dita}                                  & 84.2 & 96.3 & 85.4 & 63.8 & 82.40 \\
TraceVLA~\citep{zheng2024tracevla}                        & 84.6 & 85.2 & 75.1 & 54.1 & 74.80 \\
SpatialVLA~\citep{qu2025spatialvla}                       & 88.2 & 89.9 & 78.6 & 55.5 & 78.10 \\
$\pi_0$ FAST \citep{pertsch2025pi0fast}                   & 96.4 & 96.8 & 88.6 & 60.2 & 85.50 \\
$\pi_0$~\citep{black2024pi_0}                             & 96.8 & 98.8 & 95.8 & 85.2 & 94.15 \\
\rowcolor{gray!20} $\pi_0$ + RAS                          & 91.4 & 97.4 & 92.2 & 81.6 & 90.65 \\
\rowcolor{gray!20} $\pi_0$ + MCSI                      & 96.6 & 99.8 & 94.8 & 87.0 & 94.55 \\
\rowcolor{gray!20} $\pi_0$ + RAS \& MCSI               & 97.4 & 98.8 & 93.4 & 83.8 & 93.35 \\\midrule
$\pi_{0.5}$~\citep{intelligence2025pi05}                  & 95.4 & 98.4 & 97.2 & 89.6 & 95.15 \\
\rowcolor{gray!20} $\pi_{0.5}$ + RAS                      & 97.8 & 99.4 & 96.6 & 92.8 & 96.65 \\
\rowcolor{gray!20} $\pi_{0.5}$ + MCSI                  & 99.8 & 99.6 & 99.6 & 94.0 & 98.25 \\
\rowcolor{gray!20} $\pi_{0.5}$ + RAS \& MCSI           & 98.6 & 99.2 & 95.8 & 92.8 & 96.60 \\
\bottomrule
\end{tabular}}
\end{table*}

\begin{table*}[htbp]
\centering
\caption{
\textbf{LIBERO (blank instruction) task performance (success rate, SR) results.} In this variant, task instructions are fully blanked to remove explicit linguistic guidance.
\textbf{RAS}: Residual Affordance Steering; \textbf{MCSI}: Monte Carlo Syntactic Integration.
}
\label{tab:libero_blank}
\resizebox{0.8\textwidth}{!}{
\begin{tabular}{c|cccc|c}
\toprule
\multirow{2}{*}{\textbf{Model}} & \textbf{Libero-Spatial} & \textbf{Libero-Object} & \textbf{Libero-Goal} & \textbf{Libero-Long} & \textbf{Average} \\
& SR (\%) $\uparrow$ & SR (\%) $\uparrow$ & SR (\%) $\uparrow$ & SR (\%) $\uparrow$ & SR (\%) $\uparrow$ \\
\midrule
$\pi_0$~\citep{black2024pi_0}                             & 29.8 & 41.2 & 4.6 & 25.2 & 25.20 \\
\rowcolor{gray!20} $\pi_0$ + RAS                          & 77.4 & 95.4 & 6.8 & 74.0 & 63.40 \\
\rowcolor{gray!20} $\pi_0$ + MCSI                      & 57.0 & 50.8 & 9.8 & 49.8 & 41.85 \\
\rowcolor{gray!20} $\pi_0$ + RAS \& MCSI               & 88.0 & 98.8 & 9.8 & 82.0 & 69.65 \\\midrule
$\pi_{0.5}$~\citep{intelligence2025pi05}                  & 65.0 & 51.8 & 11.4 & 72.0 & 50.05 \\
\rowcolor{gray!20} $\pi_{0.5}$ + RAS                      & 87.0 & 99.4 & 10.8 & 84.8 & 70.50 \\
\rowcolor{gray!20} $\pi_{0.5}$ + MCSI                  & 43.4 & 50.2 & 13.8 & 77.4 & 46.20 \\
\rowcolor{gray!20} $\pi_{0.5}$ + RAS \& MCSI           & 88.4 & 99.4 & 10.2 & 83.0 & 70.25 \\
\bottomrule
\end{tabular}}
\end{table*}

\begin{table*}[htbp]
\centering
\caption{
\textbf{LIBERO (simple-word instruction) task performance (success rate, SR) results.} All task instructions are replaced by a single generic command (e.g., ”do something“), completely removing the original semantic content.
\textbf{RAS}: Residual Affordance Steering; \textbf{MCSI}: Monte Carlo Syntactic Integration.}
\label{tab:libero_simple}
\resizebox{0.8\textwidth}{!}{
\begin{tabular}{c|cccc|c}
\toprule
\multirow{2}{*}{\textbf{Model}} & \textbf{Libero-Spatial} & \textbf{Libero-Object} & \textbf{Libero-Goal} & \textbf{Libero-Long} & \textbf{Average} \\
& SR (\%) $\uparrow$ & SR (\%) $\uparrow$ & SR (\%) $\uparrow$ & SR (\%) $\uparrow$ & SR (\%) $\uparrow$ \\
\midrule
$\pi_0$~\citep{black2024pi_0}                             & 29.2 & 43.4 & 2.2 & 30.2 & 26.25 \\
\rowcolor{gray!20} $\pi_0$ + RAS                          & 79.0 & 96.2 & 5.4 & 69.4 & 62.50 \\
\rowcolor{gray!20} $\pi_0$ + MCSI                      & 58.6 & 48.8 & 9.0 & 43.0 & 39.85 \\
\rowcolor{gray!20} $\pi_0$ + RAS \& MCSI               & 86.6 & 93.4 & 9.4 & 70.2 & 64.90 \\\midrule
$\pi_{0.5}$~\citep{intelligence2025pi05}                  & 47.8 & 56.2 & 10.8 & 72.8 & 46.90 \\
\rowcolor{gray!20} $\pi_{0.5}$ + RAS                      & 86.6 & 99.2 & 9.2 & 81.4 & 69.10 \\
\rowcolor{gray!20} $\pi_{0.5}$ + MCSI                  & 43.2 & 48.2 & 15.0 & 78.4 & 46.20 \\
\rowcolor{gray!20} $\pi_{0.5}$ + RAS \& MCSI           & 88.0 & 100.0 & 10.0 & 84.4 & 70.60 \\
\bottomrule
\end{tabular}}
\end{table*}

\begin{table*}[htbp]
\centering
\caption{
\textbf{LIBERO (multi-word instruction) task performance (success rate, SR) results.} 
Instructions are rewritten by replacing words in the original sentence with simple lexical alternatives, while largely preserving the overall task semantics.
\textbf{RAS}: Residual Affordance Steering; \textbf{MCSI}: Monte Carlo Syntactic Integration.
}
\label{tab:libero_multi}
\resizebox{0.8\textwidth}{!}{
\begin{tabular}{c|cccc|c}
\toprule
\multirow{2}{*}{\textbf{Model}} & \textbf{Libero-Spatial} & \textbf{Libero-Object} & \textbf{Libero-Goal} & \textbf{Libero-Long} & \textbf{Average} \\
& SR (\%) $\uparrow$ & SR (\%) $\uparrow$ & SR (\%) $\uparrow$ & SR (\%) $\uparrow$ & SR (\%) $\uparrow$ \\
\midrule
$\pi_0$~\citep{black2024pi_0}                             & 88.6 & 97.6 & 91.4 & 87.6 & 91.30 \\

\rowcolor{gray!20} $\pi_0$ + RAS                          & 91.2 & 98.4 & 90.0 & 80.0 & 89.90 \\
\rowcolor{gray!20} $\pi_0$ + MCSI                      & 96.0 & 99.8 & 92.6 & 85.4 & 93.45 \\
\rowcolor{gray!20} $\pi_0$ + RAS \& MCSI               & 97.2 & 99.0 & 88.4 & 80.8 & 91.35 \\\midrule
$\pi_{0.5}$~\citep{intelligence2025pi05}                  & 95.2 & 99.0 & 95.0 & 92.6 & 95.45 \\
\rowcolor{gray!20} $\pi_{0.5}$ + RAS                      & 98.4 & 98.6 & 97.0 & 93.2 & 96.80 \\
\rowcolor{gray!20} $\pi_{0.5}$ + MCSI                  & 99.8 & 99.8 & 97.0 & 95.4 & 98.00 \\
\rowcolor{gray!20} $\pi_{0.5}$ + RAS \& MCSI           & 97.6 & 99.0 & 97.6 & 95.8 & 97.50 \\
\bottomrule
\end{tabular}}
\end{table*}

\begin{table*}[htbp]
\centering
\caption{
\textbf{LIBERO (random-lang instruction) task performance (success rate, SR) results.} 
Instructions are constructed by randomly shuffling the word order of the original sentence, disrupting syntactic structure while retaining the same lexical content.
\textbf{RAS}: Residual Affordance Steering; \textbf{MCSI}: Monte Carlo Syntactic Integration.
}
\label{tab:libero_random}
\resizebox{0.8\textwidth}{!}{
\begin{tabular}{c|cccc|c}
\toprule
\multirow{2}{*}{\textbf{Model}} & \textbf{Libero-Spatial} & \textbf{Libero-Object} & \textbf{Libero-Goal} & \textbf{Libero-Long} & \textbf{Average} \\
& SR (\%) $\uparrow$ & SR (\%) $\uparrow$ & SR (\%) $\uparrow$ & SR (\%) $\uparrow$ & SR (\%) $\uparrow$ \\
\midrule
$\pi_0$~\citep{black2024pi_0}                             & 83.0 & 97.4 & 91.6 & 85.4 & 89.35 \\
\rowcolor{gray!20} $\pi_0$ + RAS                          & 91.6 & 94.6 & 89.8 & 78.8 & 88.70 \\
\rowcolor{gray!20} $\pi_0$ + MCSI                      & 94.2 & 99.2 & 97.0 & 81.4 & 92.95 \\
\rowcolor{gray!20} $\pi_0$ + RAS \& MCSI               & 96.8 & 99.8 & 95.4 & 87.8 & 94.95 \\\midrule
$\pi_{0.5}$~\citep{intelligence2025pi05}                  & 96.4 & 98.8 & 94.8 & 90.8 & 95.20 \\
\rowcolor{gray!20} $\pi_{0.5}$ + RAS                      & 97.4 & 99.0 & 93.2 & 94.2 & 95.95 \\
\rowcolor{gray!20} $\pi_{0.5}$ + MCSI                  & 98.2 & 100.0 & 95.6 & 96.0 & 97.45 \\
\rowcolor{gray!20} $\pi_{0.5}$ + RAS \& MCSI           & 98.6 & 99.8 & 97.0 & 94.4 & 97.45 \\
\bottomrule
\end{tabular}}
\end{table*}

\begin{table*}[htbp]
\centering
\caption{
\textbf{LIBERO (random-mask instruction) task performance (success rate, SR) results.} 
Each word in the instruction is independently masked with a predefined probability, introducing controlled semantic degradation while preserving the original word order.
\textbf{RAS}: Residual Affordance Steering; \textbf{MCSI}: Monte Carlo Syntactic Integration.
}
\label{tab:libero_mask}

\resizebox{0.8\textwidth}{!}{
\begin{tabular}{c|cccc|c}
\toprule
\multirow{2}{*}{\textbf{Model}} & \textbf{Libero-Spatial} & \textbf{Libero-Object} & \textbf{Libero-Goal} & \textbf{Libero-Long} & \textbf{Average} \\
& SR (\%) $\uparrow$ & SR (\%) $\uparrow$ & SR (\%) $\uparrow$ & SR (\%) $\uparrow$ & SR (\%) $\uparrow$ \\
\midrule

\multicolumn{6}{c}{\textbf{Random Mask Rate = 0.2}} \\
\midrule
$\pi_0$~\citep{black2024pi_0}                             & 53.4 & 90.2 & 74.8 & 72.2 & 72.65 \\
\rowcolor{gray!30} $\pi_0$ + RAS                          & 69.6 & 94.6 & 75.4 & 74.2 & 78.45 \\
\rowcolor{gray!30} $\pi_0$ + MCSI                      & 94.2 & 97.4 & 81.0 & 82.8 & 88.85 \\
\rowcolor{gray!30} $\pi_0$ + RAS \& MCSI               & 96.4 & 99.8 & 92.6 & 79.4 & 92.05 \\\midrule
$\pi_{0.5}$~\citep{intelligence2025pi05}                  & 95.0 & 97.4 & 89.6 & 88.6 & 92.65 \\
\rowcolor{gray!30} $\pi_{0.5}$ + RAS                      & 96.0 & 98.6 & 88.2 & 92.8 & 93.90 \\
\rowcolor{gray!30} $\pi_{0.5}$ + MCSI                  & 98.4 & 99.2 & 91.8 & 94.8 & 96.05 \\
\rowcolor{gray!30} $\pi_{0.5}$ + RAS \& MCSI           & 98.4 & 99.6 & 92.0 & 95.4 & 96.35 \\
\midrule

\multicolumn{6}{c}{\textbf{Random Mask Rate = 0.4}} \\
\midrule
$\pi_0$~\citep{black2024pi_0}                             & 15.2 & 75.0 & 33.0 & 46.8 & 42.50 \\
\rowcolor{gray!30} $\pi_0$ + RAS                          & 34.2 & 83.8 & 44.8 & 57.4 & 55.05 \\
\rowcolor{gray!30} $\pi_0$ + MCSI                      & 87.8 & 93.6 & 51.8 & 77.8 & 77.75 \\
\rowcolor{gray!30} $\pi_0$ + RAS \& MCSI               & 93.6 & 100.0 & 71.8 & 78.0 & 85.85 \\\midrule
$\pi_{0.5}$~\citep{intelligence2025pi05}                  & 87.2 & 90.8 & 67.6 & 85.2 & 82.70 \\
\rowcolor{gray!30} $\pi_{0.5}$ + RAS                      & 89.6 & 98.8 & 64.6 & 93.6 & 86.65 \\
\rowcolor{gray!30} $\pi_{0.5}$ + MCSI                  & 87.6 & 95.4 & 72.0 & 93.8 & 87.20 \\
\rowcolor{gray!30} $\pi_{0.5}$ + RAS \& MCSI           & 97.0 & 100.0 & 75.6 & 95.4 & 92.00 \\
\midrule

\multicolumn{6}{c}{\textbf{Random Mask Rate = 0.6}} \\
\midrule
$\pi_0$~\citep{black2024pi_0}                             & 7.4 & 51.0 & 10.0 & 20.2 & 22.15 \\
\rowcolor{gray!30} $\pi_0$ + RAS                          & 18.2 & 66.8 & 20.4 & 28.8 & 33.55 \\
\rowcolor{gray!30} $\pi_0$ + MCSI                      & 81.4 & 84.0 & 24.8 & 65.2 & 63.85 \\
\rowcolor{gray!30} $\pi_0$ + RAS \& MCSI               & 92.2 & 97.4 & 50.6 & 71.6 & 77.95 \\\midrule
$\pi_{0.5}$~\citep{intelligence2025pi05}                  & 71.8 & 79.2 & 43.4 & 83.8 & 69.55 \\
\rowcolor{gray!30} $\pi_{0.5}$ + RAS                      & 87.4 & 99.4 & 44.0 & 86.4 & 79.30 \\
\rowcolor{gray!30} $\pi_{0.5}$ + MCSI                  & 70.8 & 85.2 & 46.6 & 91.0 & 73.40 \\
\rowcolor{gray!30} $\pi_{0.5}$ + RAS \& MCSI           & 93.6 & 100.0 & 50.6 & 94.0 & 84.55 \\
\midrule

\multicolumn{6}{c}{\textbf{Random Mask Rate = 0.8}} \\
\midrule
$\pi_0$~\citep{black2024pi_0}                             & 0.8  & 13.6 & 5.4  & 11.4 & 7.80 \\
\rowcolor{gray!30} $\pi_0$ + RAS                          & 9.0  & 36.0 & 8.8  & 18.0 & 17.95 \\
\rowcolor{gray!30} $\pi_0$ + MCSI                      & 73.4 & 69.2 & 12.8 & 55.8 & 52.80 \\
\rowcolor{gray!30} $\pi_0$ + RAS \& MCSI               & 89.0 & 95.0 & 29.2 & 66.4 & 69.90 \\\midrule
$\pi_{0.5}$~\citep{intelligence2025pi05}                  & 60.0 & 63.2 & 24.6 & 74.0 & 55.45 \\
\rowcolor{gray!30} $\pi_{0.5}$ + RAS                      & 83.0 & 97.4 & 23.0 & 80.8 & 71.05 \\
\rowcolor{gray!30} $\pi_{0.5}$ + MCSI                  & 50.8 & 69.4 & 27.0 & 84.6 & 57.95 \\
\rowcolor{gray!30} $\pi_{0.5}$ + RAS \& MCSI           & 89.0 & 99.6 & 29.8 & 91.6 & 77.50 \\

\bottomrule
\end{tabular}}
\end{table*}

\section{Detailed Ablation Results} 
\label{appendix:detailed_ablation}

Tables~\ref{tab:ablation_sc}, \ref{tab:ablation_steps}, and~\ref{tab:ablation_steps_pi05} present detailed ablation results on the steering coefficient (SC) and the number of denoising steps under destructive instruction overwriting. All results are reported as average success rates across different language perturbation benchmarks.

As shown in Table~\ref{tab:ablation_sc}, the steering coefficient plays a critical role in balancing robustness and over-steering. Excessively large coefficients lead to noticeable performance degradation. This indicates that overly strong residual affordance steering can dominate the policy update, thereby reducing adaptability under corrupted language inputs.

Tables~\ref{tab:ablation_steps} and~\ref{tab:ablation_steps_pi05} further show the effect of denoising steps for $\pi_0$ and $\pi_{0.5}$, respectively. Notably, $\pi_{0.5}$ exhibits stronger overall stability across different step configurations, indicating that a stronger base language grounding reduces sensitivity to inference-time hyperparameters.

Overall, these results demonstrate that robustness under destructive instruction overwriting depends on a careful balance between steering strength and inference depth. Moderate steering coefficients combined with sufficient  denoising steps yield the most stable and consistent performance across language perturbations.

\begin{table*}[htbp]
\centering
\caption{
\textbf{Ablation on steering coefficient (SC) across different language perturbation benchmarks.} Denoising steps are fixed to 10. All values are average success rate (SR, \%).
}
\label{tab:ablation_sc}
\resizebox{0.95\textwidth}{!}{
\begin{tabular}{c|ccccccccc|c}
\toprule
\multirow{2}{*}{\textbf{SC}} 
& \textbf{Origin}
& \textbf{Multi}
& \textbf{Blank}
& \textbf{Rand}
& \textbf{M2}
& \textbf{M4}
& \textbf{M6}
& \textbf{M8}
& \textbf{Simple}
& \textbf{Average} \\
& SR (\%) $\uparrow$ & SR (\%) $\uparrow$ & SR (\%) $\uparrow$ & SR (\%) $\uparrow$ & SR (\%) $\uparrow$ & SR (\%) $\uparrow$ & SR (\%) $\uparrow$ & SR (\%) $\uparrow$ & SR (\%) $\uparrow$ & SR (\%) $\uparrow$ \\

\midrule
\multicolumn{11}{c}{$\pi_0$~\citep{black2024pi_0} + RAS} \\
\midrule
1.25                     & 92.45 & 91.00 & 63.25 & 91.05 & 80.40 & 61.00 & 39.45 & 24.15 & 63.40 & 67.35 \\
1.5                      & 90.65 & 89.90 & 63.40 & 88.70 & 78.45 & 55.05 & 33.55 & 17.95 & 62.50 & 64.46 \\
2.0                      & 89.50 & 87.95 & 66.15 & 87.90 & 71.75 & 45.45 & 22.05 & 9.00  & 62.05 & 60.20 \\
3.0                      & 81.95 & 82.15 & 64.00 & 79.20 & 58.00 & 25.80 & 7.90  & 1.90  & 58.80 & 51.08 \\

\midrule
\multicolumn{11}{c}{$\pi_{0.5}$~\citep{intelligence2025pi05}  + RAS} \\
\midrule
1.05                     & 97.50 & 96.70 & 69.95 & 96.75 & 93.05 & 84.30 & 77.95 & 71.80 & 70.55 & 84.28 \\
1.1                      & 97.25 & 97.05 & 70.35 & 96.15 & 93.25 & 84.50 & 77.30 & 72.30 & 69.80 & 84.22 \\
1.2                      & 97.50 & 97.00 & 69.25 & 95.85 & 93.50 & 85.20 & 79.05 & 72.40 & 69.90 & 84.41 \\
1.25                     & 97.40 & 96.80 & 70.00 & 96.55 & 94.70 & 86.85 & 78.30 & 72.20 & 70.10 & \textbf{84.77} \\
1.5                      & 96.65 & 96.80 & 70.50 & 95.95 & 93.90 & 86.65 & 79.30 & 71.05 & 69.10 & 84.43 \\
1.75                     & 96.90 & 96.50 & 70.00 & 96.10 & 94.85 & 85.90 & 77.75 & 69.20 & 69.75 & 84.11 \\
2.0                      & 96.70 & 97.00 & 69.45 & 96.25 & 94.00 & 86.00 & 77.40 & 66.90 & 69.15 & 83.65 \\
3.0                      & 96.45 & 96.90 & 68.80 & 96.60 & 94.75 & 87.60 & 75.35 & 55.65 & 68.55 & 82.29 \\

\bottomrule
\end{tabular}}
\end{table*}

\begin{table*}[htbp]
\centering
\caption{
\textbf{Ablation on denoising steps across different \textit{destructive instruction overwriting} benchmarks.} The base model is $\pi_{0}$~\citep{black2024pi_0}. The steering coefficient is fixed to 1.5. All values are average success rate (SR, \%).
}
\label{tab:ablation_steps}
\resizebox{0.95\textwidth}{!}{
\begin{tabular}{c|ccccccccc|c}
\toprule
\multirow{2}{*}{\textbf{Steps}} 
& \textbf{Origin}
& \textbf{Multi}
& \textbf{Blank}
& \textbf{Rand}
& \textbf{M2}
& \textbf{M4}
& \textbf{M6}
& \textbf{M8}
& \textbf{Simple}
& \textbf{Average} \\
& SR (\%) $\uparrow$ & SR (\%) $\uparrow$ & SR (\%) $\uparrow$ & SR (\%) $\uparrow$ & SR (\%) $\uparrow$ & SR (\%) $\uparrow$ & SR (\%) $\uparrow$ & SR (\%) $\uparrow$ & SR (\%) $\uparrow$ & SR (\%) $\uparrow$ \\
\midrule
5                       & 93.55 & 92.50 & 66.10 & 91.80 & 80.65 & 56.15 & 34.95 & 18.65 & 65.00 & 66.59 \\
10                      & 90.65 & 89.90 & 63.40 & 88.80 & 78.45 & 55.05 & 33.55 & 17.95 & 62.50 & 64.46 \\
15                      & 90.60 & 89.05 & 63.50 & 89.05 & 75.15 & 53.05 & 32.25 & 16.90 & 62.55 & 63.57 \\
\bottomrule
\end{tabular}}
\end{table*}

\begin{table*}[htbp]
\centering
\caption{
\textbf{Ablation on denoising steps across different \textit{destructive instruction overwriting} benchmarks.} The base model is $\pi_{0.5}$~\citep{intelligence2025pi05}. The steering coefficient is fixed to 1.25. All values are average success rate (SR, \%).
}
\label{tab:ablation_steps_pi05}
\resizebox{0.95\textwidth}{!}{
\begin{tabular}{c|ccccccccc|c}
\toprule
\multirow{2}{*}{\textbf{Steps}} 
& \textbf{Origin}
& \textbf{Multi}
& \textbf{Blank}
& \textbf{Rand}
& \textbf{M2}
& \textbf{M4}
& \textbf{M6}
& \textbf{M8}
& \textbf{Simple}
& \textbf{Average} \\
& SR (\%) $\uparrow$ & SR (\%) $\uparrow$ & SR (\%) $\uparrow$ & SR (\%) $\uparrow$ & SR (\%) $\uparrow$ & SR (\%) $\uparrow$ & SR (\%) $\uparrow$ & SR (\%) $\uparrow$ & SR (\%) $\uparrow$ & SR (\%) $\uparrow$ \\
\midrule
5                       & 96.90 & 97.35 & 69.90 & 96.65 & 93.90 & 86.50 & 78.70 & 71.45 & 68.90 & 84.47 \\
15                      & 97.40 & 97.20 & 69.35 & 95.60 & 94.30 & 85.00 & 78.90 & 72.00 & 69.60 & 84.37 \\
20                      & 96.85 & 96.60 & 70.30 & 96.65 & 93.45 & 85.60 & 78.05 & 71.60 & 69.80 & 84.32 \\
\bottomrule
\end{tabular}}
\end{table*}

\section{Qualitative Analysis under Destructive Instruction Overwriting}
\label{appendix:qualitative_analysis}

\begin{figure*}[t]
  \centering
  \includegraphics[width=0.6\textwidth]{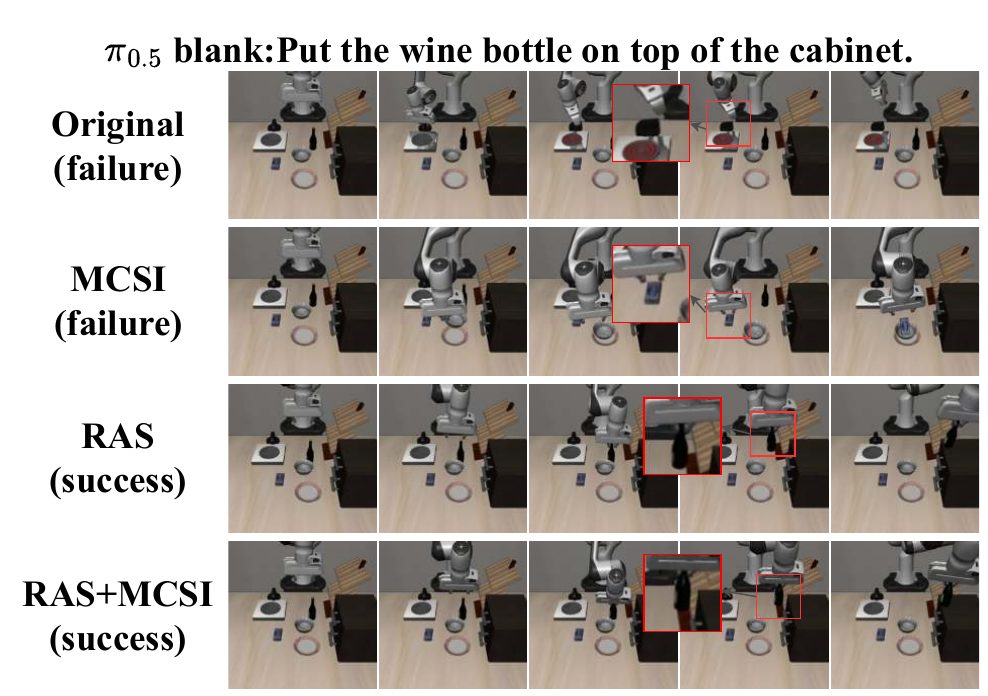}
  \caption{
  \textbf{Qualitative comparison under destructive instruction overwriting ($\pi_{0.5}$).}
    We visualize representative rollout trajectories for the task “Put the wine bottle on top of the cabinet” when the instruction is partially blanked. The base model is $\pi_{0.5}$~\citep{intelligence2025pi05}.
    \textbf{RAS}: Residual Affordance Steering; \textbf{MCSI}: Monte Carlo Syntactic Integration.
  }
  \label{fig:pi05_blank_compare}
\end{figure*}

\begin{figure*}[t]
  \centering
  \includegraphics[width=0.6\textwidth]{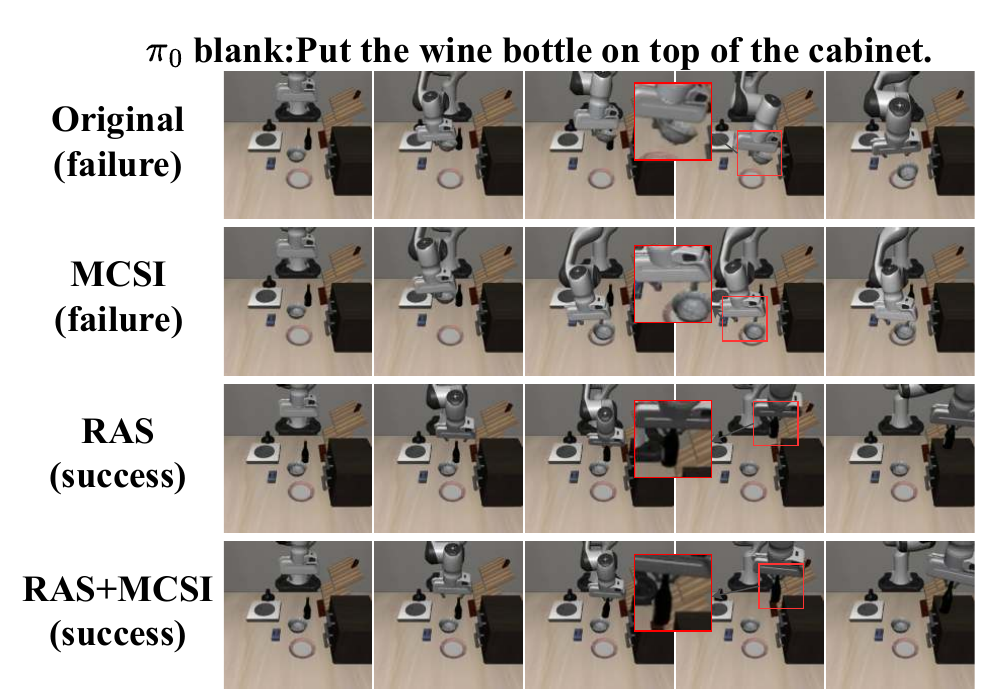}
  \caption{
  \textbf{Qualitative comparison under destructive instruction overwriting ($\pi_{0}$).}
    We visualize representative rollout trajectories for the task “Put the wine bottle on top of the cabinet” when the instruction is partially blanked. The base model is $\pi_{0.5}$~\citep{black2024pi_0}.
    \textbf{RAS}: Residual Affordance Steering; \textbf{MCSI}: Monte Carlo Syntactic Integration.
  }
  \label{fig:pi0_blank_compare}
\end{figure*}

Figures~\ref{fig:pi05_blank_compare} and~\ref{fig:pi0_blank_compare} present qualitative rollouts under \emph{destructive instruction overwriting}, where key linguistic tokens are removed from the command.
In this setting, the vanilla policies exhibit brittle behavior: although the scene remains visually unchanged, the policy fails to consistently localize the wine bottle or align it with the cabinet-top affordance, resulting in unstable grasps or premature terminations.

Applying \textbf{MCSI} improves robustness to syntactic variation by aggregating multiple rewritten instructions; however, when critical semantic content is missing, MCSI alone is insufficient to recover the correct action intent.
In contrast, \textbf{RAS} explicitly injects residual affordance signals derived from visual observations, steering the policy toward task-relevant object--target configurations even when language supervision is severely degraded.

Notably, the combination of \textbf{RAS+MCSI} achieves the most reliable performance.
By jointly mitigating linguistic uncertainty and reinforcing visual affordance alignment, the policy consistently executes the correct placement behavior across all evaluated rollouts.
This result highlights that robustness to destructive language perturbations requires not only syntactic integration but also explicit affordance-level correction.

\section{Instruction Rewriting Examples on Obfuscated Instruction Reinterpretation} \label{appendix:instruction_rewrite}

\begin{figure*}[t]
  \centering
  \includegraphics[width=0.8\textwidth]{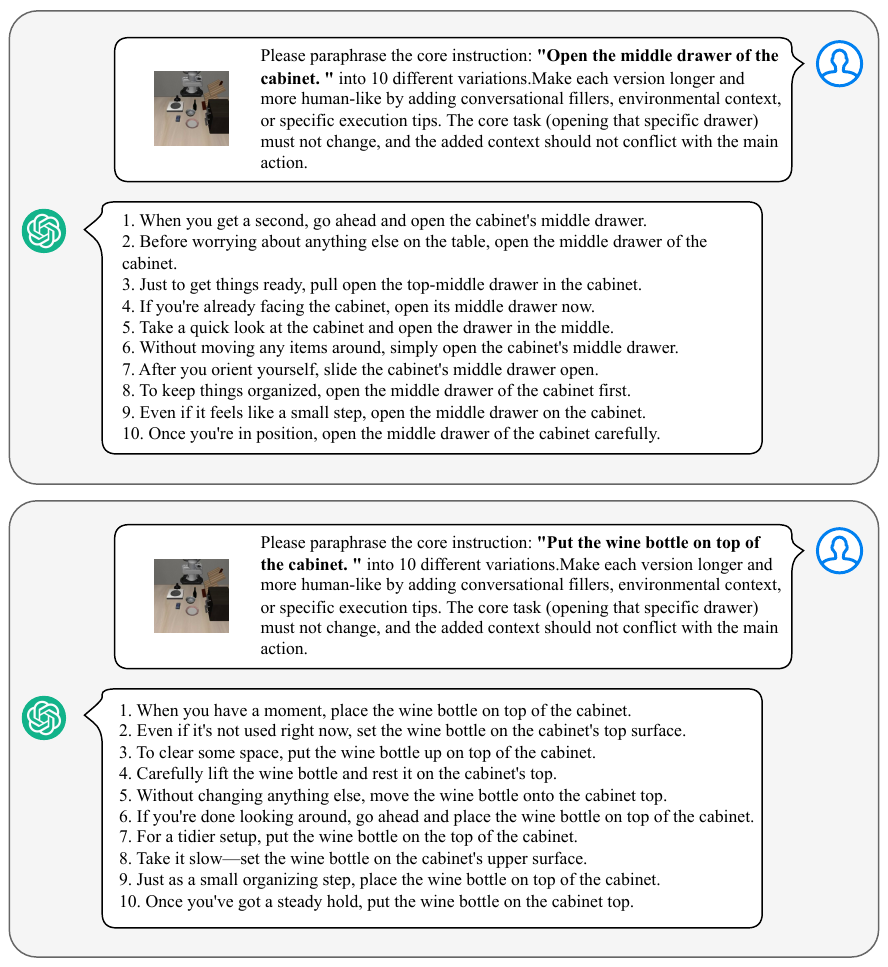}
  \caption{
  \textbf{R1-Distraction.}
  The instruction is augmented with task-irrelevant conversational or contextual content, such as background descriptions or auxiliary remarks, while keeping the core action and target unchanged.
  }
  \label{fig:r1_dialog}
\end{figure*}

\begin{figure*}[t]
  \centering
  \includegraphics[width=0.8\textwidth]{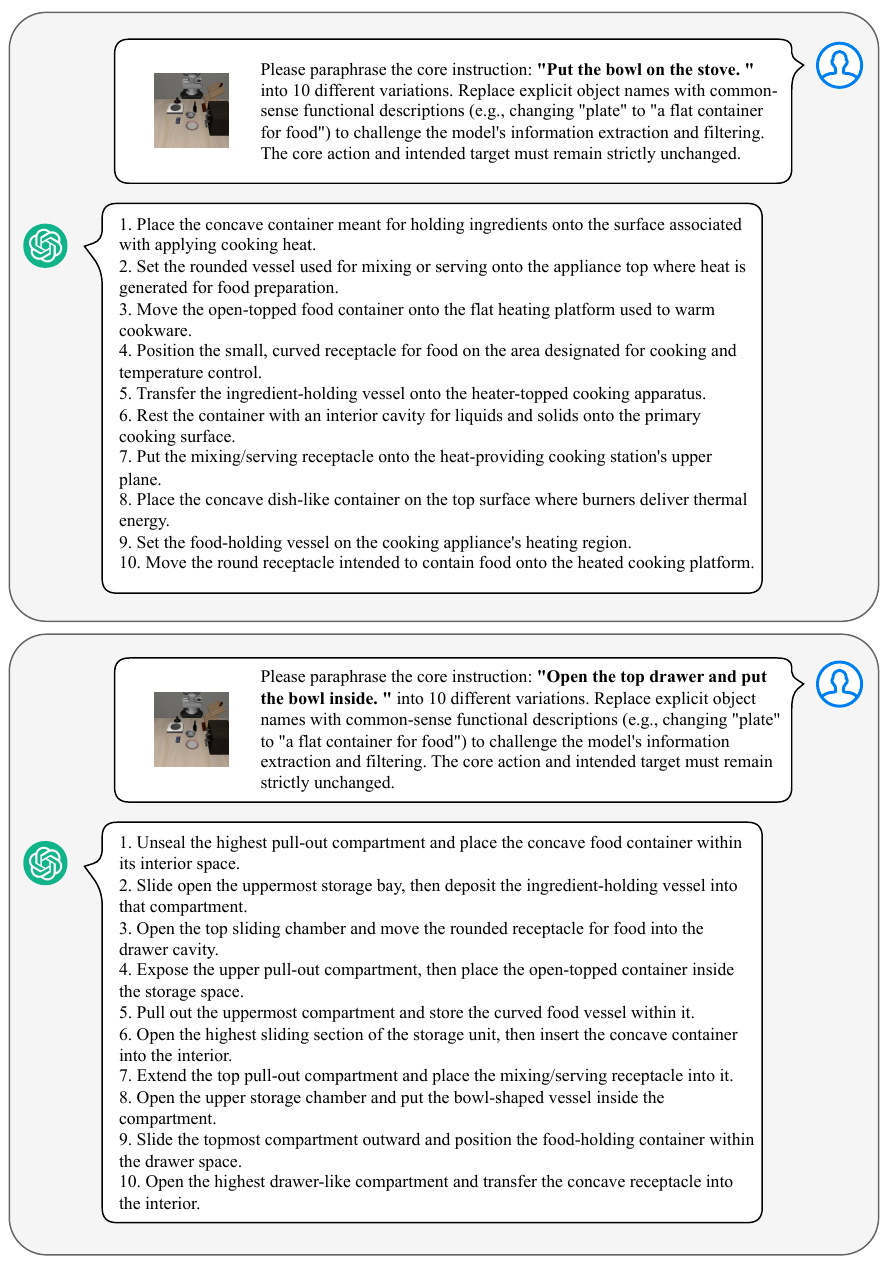}
  \caption{
  \textbf{R2-Common Sense.}
  Object names are replaced with commonsense-based descriptive phrases that implicitly convey their functional or physical properties. Although the task intent remains unchanged, this variant requires the model to extract relevant semantics from more abstract and verbose descriptions.
  }
  \label{fig:r2_dialog}
\end{figure*}

\begin{figure*}[t]
  \centering
  \includegraphics[width=0.8\textwidth]{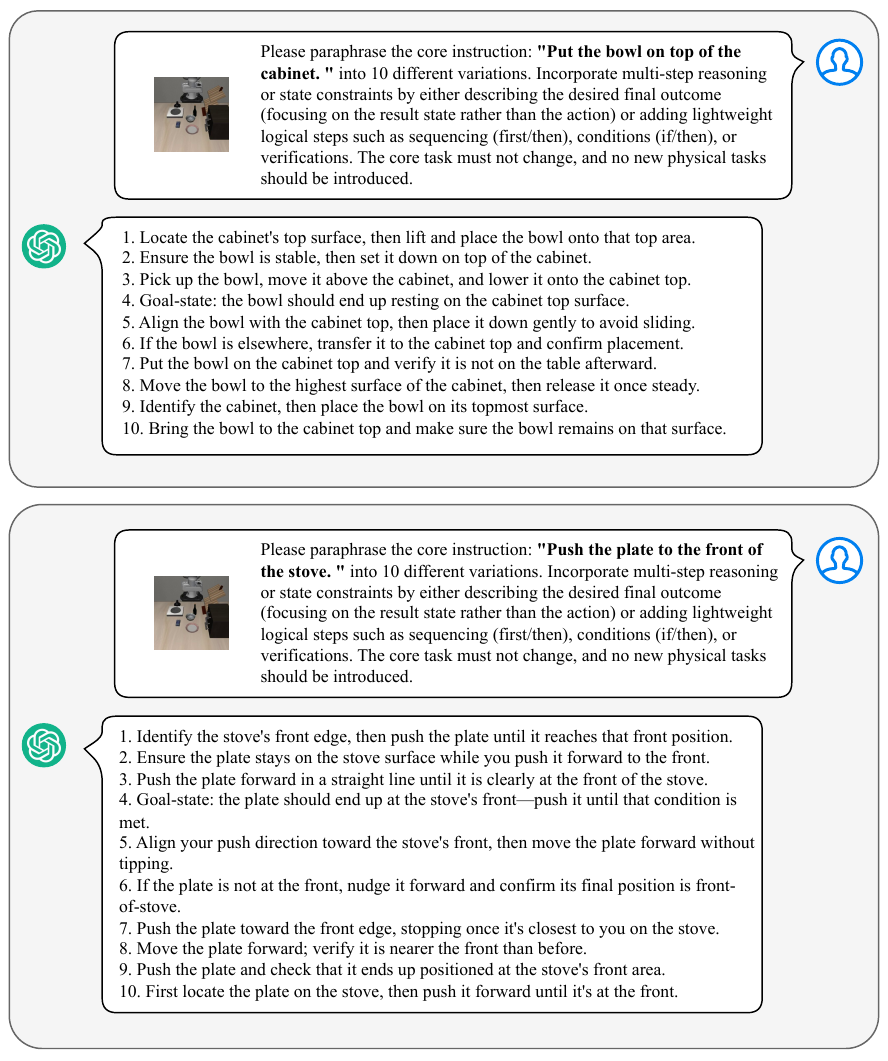}
  \caption{
  \textbf{R3-Reasoning Chain.}
  The instruction is reformulated to emphasize implicit reasoning, execution order, or final-state constraints, either by introducing lightweight reasoning cues or by abstracting intermediate steps. The target task remains identical, but the linguistic form encourages reasoning-based interpretation.
  }
  \label{fig:r3_dialog}
\end{figure*}

\begin{figure*}[t]
  \centering
  \includegraphics[width=0.8\textwidth]{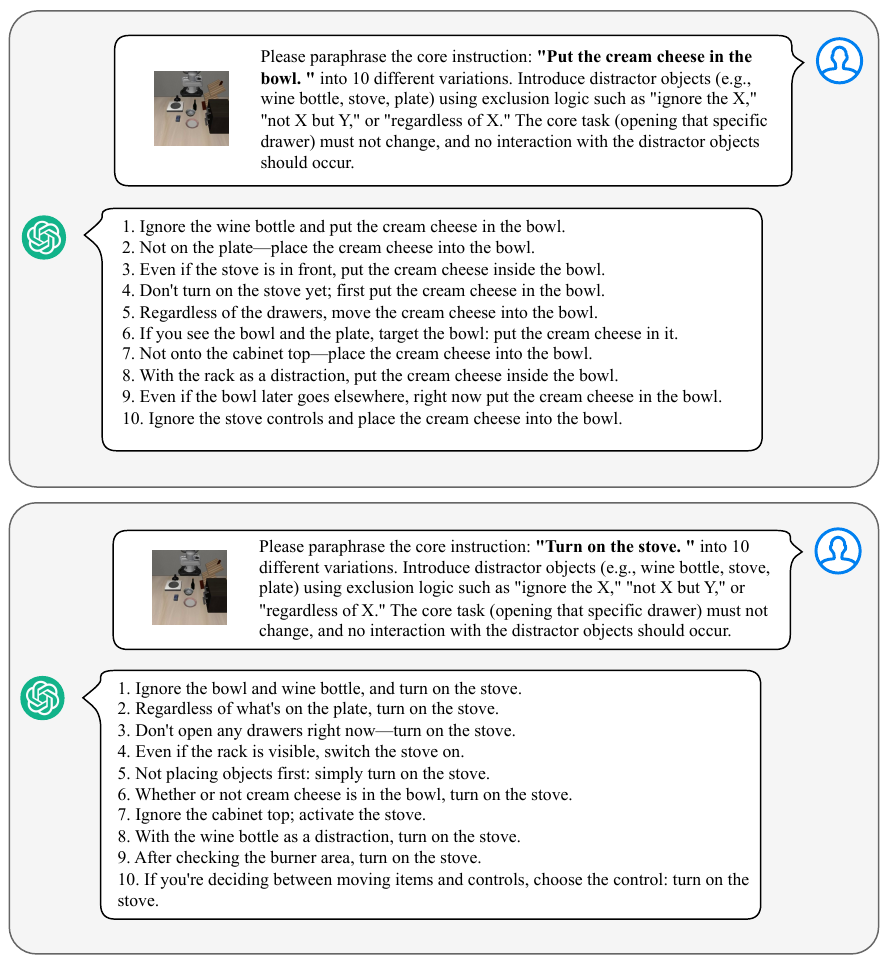}
  \caption{
  \textbf{R4-Confusion.}
  The instruction explicitly introduces distractor objects or actions through negation or contrast, while still specifying the correct target object and goal. This variant probes the model’s ability to resist object-level confusion and focus on task-relevant semantics.
  }
  \label{fig:r4_dialog}
\end{figure*}

As illustrated in Figures~\ref{fig:r1_dialog}, \ref{fig:r2_dialog}, \ref{fig:r3_dialog}, and \ref{fig:r4_dialog}, we employ ChatGPT-5.2 \citep{openai_chatgpt} to automatically rewrite the original task instructions into a set of semantically equivalent but linguistically obfuscated variants. These examples demonstrate how different rewriting strategies introduce increasing levels of surface variation, contextual distraction, and semantic indirection while strictly preserving the underlying task goal. Specifically, R1 and R2 introduce task-irrelevant context and commonsense-based object descriptions, respectively. R3 further reformulates instructions by emphasizing implicit reasoning or final-state constraints, and R4 explicitly injects object-level distractors via negation.

\end{document}